\pdfoutput=1

\documentclass[11pt]{article}

\usepackage[table]{xcolor}
\usepackage[]{acl}

\usepackage{url}
\makeatletter
\g@addto@macro{\UrlBreaks}{\UrlOrds}
\makeatother

\usepackage{times}
\usepackage{latexsym}

\usepackage[T1]{fontenc}

\usepackage[utf8]{inputenc}

\usepackage{mdframed}

\usepackage{inconsolata}

\usepackage[capitalize]{cleveref}
\usepackage{graphicx}
\usepackage{longtable}
\usepackage{multirow}
\usepackage{wasysym}
\usepackage{enumitem}
\usepackage{hhline}
\usepackage{makecell}
\usepackage{fontawesome5}

\usepackage{dot2texi,tikz}

\newcommand{\dataset}{\textsc{AlgoPuzzleVQA}}

\usepackage{soul,xcolor}
\usepackage{booktabs}
\usepackage{multirow}
\usepackage{pifont}
\usepackage{array} 
\usepackage{adjustbox} 
\crefformat{section}{\S#2#1#3} 
\crefformat{subsection}{\S#2#1#3}
\crefformat{subsubsection}{\S#2#1#3}

\newcommand{\code}[1]{\texttt{#1}}
\newcommand{\cmark}{\ding{51}}%
\newcommand{\xmark}{\ding{55}}%
\newcommand{\greencheck}{{\color{cyan}\cmark}}
\newcommand{\redcross}{{\color{red}\xmark}}
\let\realcite\cite
\renewcommand{\cite}[1]{\ifx.#1.\hl{[?]}\else\realcite{#1}\fi}
\newcolumntype{P}[1]{>{\raggedright\arraybackslash}p{#1}}
\newcommand{\greyrule}{\arrayrulecolor{black!30}\midrule\arrayrulecolor{black}}

\title{Visual Question Answering for Algorithmic Puzzles}

\title{\textsc{Are Language Models Puzzle Prodigies?} \\ Algorithmic Puzzles Unveil Serious Challenges in Multimodal Reasoning}

\author{Deepanway Ghosal$^1$, Vernon Toh Yan Han$^1$, Chia Yew Ken$^1$, Soujanya Poria$^1$ \\
$^1$ Singapore University of Technology and Design\\
}

\begin{document}
\maketitle
\begin{minipage}[t]{2\linewidth}
\vspace{-1.75cm}
  \centering
  \faGithub: \url{https://github.com/declare-lab/LLM-PuzzleTest} \\
  \faGlobe : \url{https://algopuzzlevqa.github.io/}
\vspace{0.5cm}
\end{minipage}

\begin{abstract}
This paper introduces the novel task of multimodal puzzle solving, framed within the context of visual question-answering. We present a new dataset, \dataset{} designed to challenge and evaluate the capabilities of multimodal language models in solving algorithmic puzzles that necessitate both visual understanding, language understanding, and complex algorithmic reasoning. We create the puzzles to encompass a diverse array of mathematical and algorithmic topics such as boolean logic, combinatorics, graph theory, optimization, search, etc., aiming to evaluate the gap between visual data interpretation and algorithmic problem-solving skills. The dataset is generated automatically from code authored by humans. All our puzzles have exact solutions that can be found from the algorithm without tedious human calculations. It ensures that our dataset can be scaled up arbitrarily in terms of reasoning complexity and dataset size. Our investigation reveals that large language models (LLMs) such as GPT4V and Gemini exhibit limited performance in puzzle-solving tasks. We find that their performance is near random in a multi-choice question-answering setup for a significant number of puzzles. The findings emphasize the challenges of integrating visual, language, and algorithmic knowledge for solving complex reasoning problems.

\end{abstract}


\begin{figure*}[t]
  \centering
  \includegraphics[width=\textwidth]{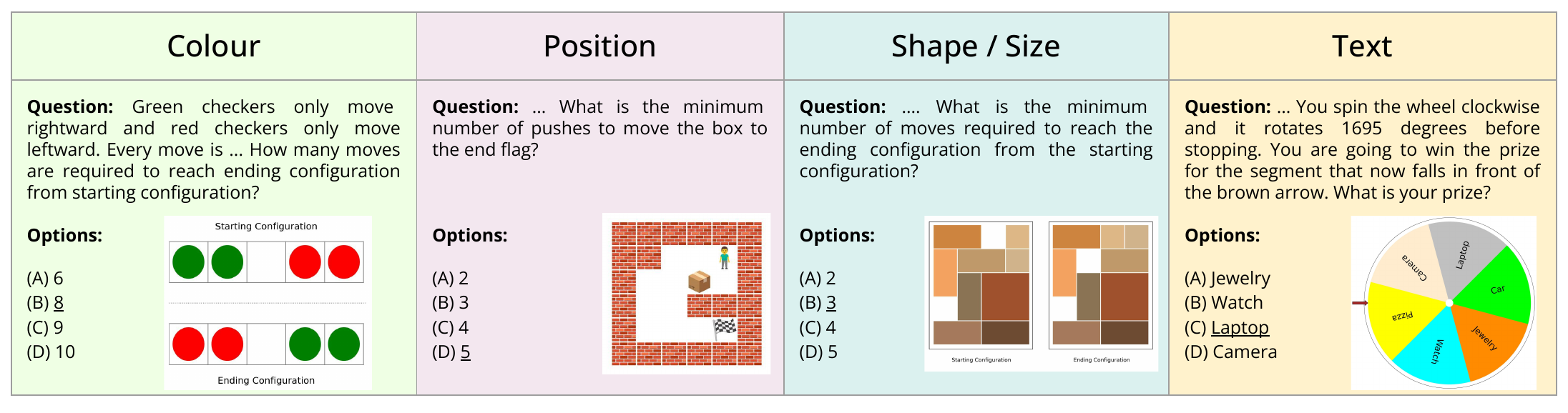}
  \caption{\footnotesize{Examples of puzzles in \dataset{} based on \textbf{visual features}. We show one instance of a puzzle from each of the four visual categories. Note that the header categories are not exhaustive as some puzzles may belong to more visual categories than the ones shown in the header. The complete ontological categorization can be found in \Cref{tab:ontology}. The questions shown above have been shortened to fit in the figure. The full questions can be found later in \Cref{sec:puzzles} and in \Cref{sec:appendix:dataset}.}}
  \label{fig:teaser_visual}
\end{figure*}

\section{Introduction}
Puzzles have long been a source of fascination and intellectual challenge, serving both as entertaining pastimes and as means to advance mathematical and logical understanding. Historically, puzzles have often emerged from complex problems, with their solutions contributing valuable insights to mathematical research~\cite{euler1741solutio,wiles1995modular,knuth2000dancing}. In this paper, we explore the novel task of algorithmic visual puzzle solving. Traditional visual question-answering (VQA)~\cite{antol2015vqa,goyal2017making} and visual reasoning~\cite{hudson2018gqa} datasets have mainly focused on combining language and vision through the lens of object detection, 
scene recognition, 
and spatial relationship, 
where the answer can be directly found in the image. Another line of work has explored the application of external factual knowledge 
~\cite{shahMYP19} or commonsense world knowledge 
~\cite{schwenk2022okvqa} for VQA. Recently, multimodal reasoning benchmarks such as ScienceQA~\cite{lu2022learn}, and MMMU~\cite{yue2023mmmu} have been proposed that aim to evaluate the expert knowledge of LLMs through subject-specific questions spanning science, medicine, business, arts, etc. 

Unlike previous research, our work integrates a new critical component -- algorithmic reasoning, which demands the ability to apply logical rules to novel situations, rather than relying on pre-stored information or direct recall of facts. This distinction is vital as it shifts the focus from merely memorizing vast databases of knowledge to applying concrete reasoning steps for novel problem-solving. We thus propose to benchmark the problem-solving ability of AI models through visual puzzles. Our puzzles are created to be self-contained where minimal additional knowledge (beyond basic mathematics)
is required to solve them. This helps us to disentangle the \textit{knowing} part from the \textit{solving} part, as the knowledge required to solve the problem is already provided as context. 


Consider \Cref{fig:teaser_visual} as an example of some of our puzzles. We show the setting or the configuration of the puzzle/problem as an image, which we consider as the \textbf{visual context}. Depending on the problem, the visual context may show different kinds of information such as the initial and final configurations of the puzzle, the position of entities on a grid, the colours of a map, etc. Visual features such as colours, positions, shapes, and sizes of the puzzle components are generally present in the visual context. The \textbf{language context}, which is shown as \textit{Question} in the figure then describes the specific features of the puzzle and the associated rules. It also presents the query problem that we are trying to solve. Correctly answering this question requires applying a fundamental \textbf{mathematical or algorithmic concept}. We design the puzzles such that solving them requires a wide array of mathematical and algorithmic topics, ranging from basic arithmetic, set theory, and boolean logic to more sophisticated areas like graph theory, and optimized search. Our proposed dataset thus offers a challenge that requires a synergistic application of language, visual, and algorithmic skills to solve complex reasoning problems.

\begin{figure*}[t]
  \centering
  \includegraphics[width=\textwidth]{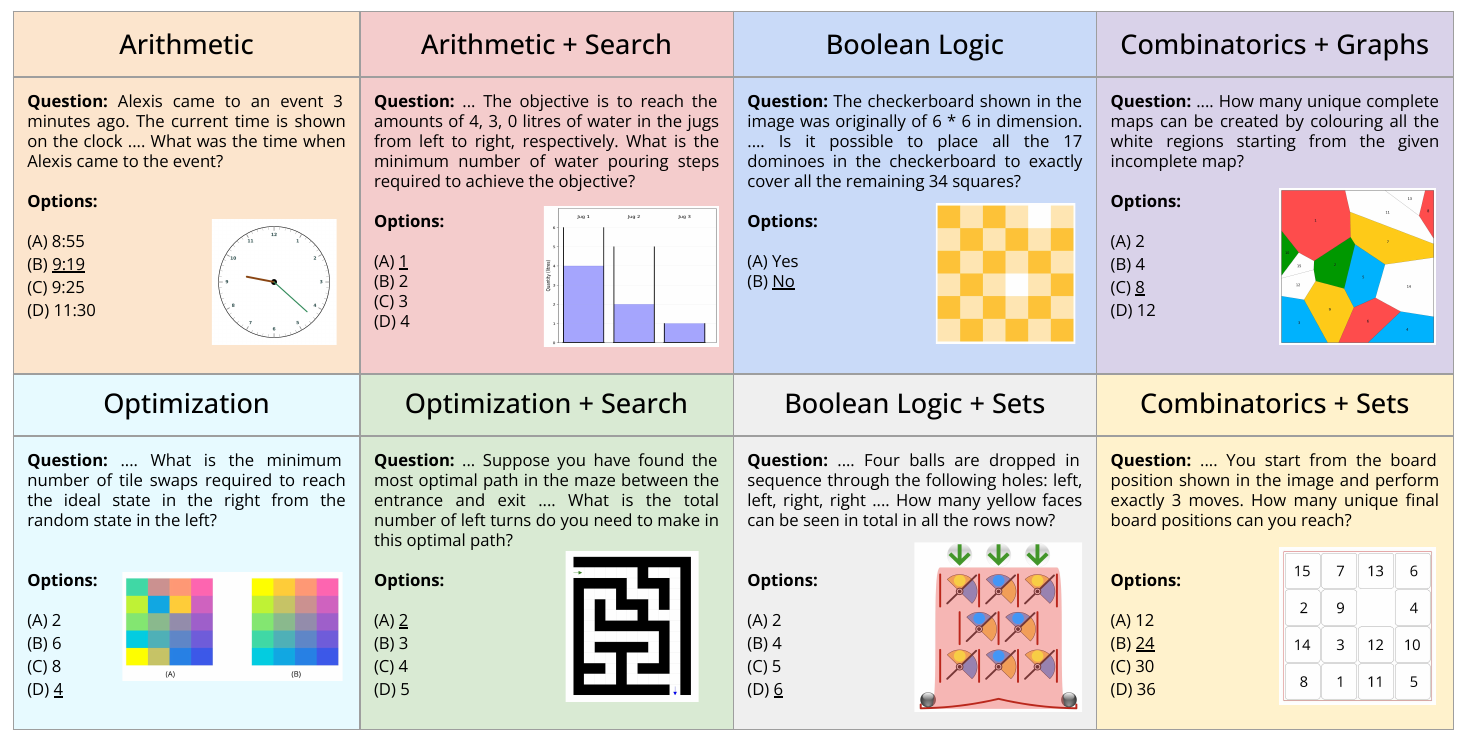}
  \caption{\footnotesize{Examples of puzzles from \dataset{} based on \textbf{algorithmic features}. We show at least one instance of a puzzle from each of the seven algorithmic categories. Note that the header categories are not exhaustive as some puzzles may belong to more algorithmic categories than the ones shown in the header.}}
  \label{fig:teaser}
\end{figure*}

A key advantage of our approach is the automated generation of puzzles from human-written code, which allows for the scalable creation of challenges with adjustable reasoning complexity. For a particular puzzle, we first write the general algorithmic solution based on well known proofs and results from the literature. By varying the inputs and configurations of the problem, we can create instances of desired difficulties. 
As multimodal language models become more capable~\cite{gpt4vision,geminipro}, it is crucial that we design clean and more challenging benchmarks to make fair assessments of their capabilities. Our automatic generation framework offers a simple way of creating and then continuously updating the benchmark with increasingly complex problems to evaluate progressively stronger multimodal models in the future. 


We also highlight another feature of our dataset -- the algorithmic foundation of our puzzle generation process ensures that each instance in our dataset has a definitive solution, eliminating the possibility of annotation error~\cite{northcutt2021pervasive}, subjectivity, ambiguities and biases that may arise from human annotated content~\cite{geva2019we}. 

We summarize our contributions as follows:
\begin{itemize}[itemsep=0pt, leftmargin=*, wide, labelwidth=0pt, labelindent=0pt, parsep=0pt]
\item We introduce an ontology of multimodal reasoning features tailored for visual algorithmic puzzle solving, aimed at delineating the capabilities and limitations of LLMs. Since each puzzle is annotated with distinct visual e.g., colour, position and algorithmic characteristics e.g., arithmetic, and search derived from our ontology, we have gained valuable insights into the specific skills that LLMs lack in addressing individual problems.
\item Relying on this ontology, we created a novel puzzle dataset designed to test multimodal reasoning across vision, language, and algorithms. All puzzles in our dataset possess unambiguous algorithmic solutions. The distinguishing feature of our proposed dataset, \dataset{}, resides in the prerequisites of the solutions to the problems. They demand proficiency in algorithmic and mathematical reasoning by applying logic adeptly in novel contexts without necessitating specialized domain knowledge.
\item Our dataset is automatically generated using a scalable puzzle generation framework, allowing for dynamic adjustment of problem complexity.
\item Experimental results conducted on this dataset with GPT-4V, Gemini Pro, and other models such as InstructBlip demonstrate challenges in achieving satisfactory scores. Our analysis reveals a fundamental deficiency in these models' visual perception and algorithmic reasoning capabilities, hindering effective complex reasoning.
\end{itemize}

\section{The Proposed Ontology}
Our puzzles encompass a diverse range of visual, mathematical and algorithmic topics. We categorize the visual and algorithmic features present in each puzzle in \Cref{tab:ontology}. 

\subsection{Visual Features}
The configuration of the puzzle/problem is shown as an image, which constitutes its visual context. We identify some fundamental aspects of the visual context that influence the nature of the puzzles. We show examples of the categories in \Cref{fig:teaser_visual}. The categorization is as follows:

\noindent{\textbf{Colour:}} The category includes puzzles where understanding the colour of the puzzle components is crucial for solving the question. For these puzzles, a different colour combination of the components in the visual context would generally lead to a different solution even for the same question. \textit{Checker Move}, \textit{Colour Hue}, \textit{Rubik's Cube}, etc. come in this category. 

\noindent{\textbf{Position:}} The puzzles where the position of the puzzle components is necessary for solving the question. All puzzles of our dataset fall under this category. Positional understanding is thus necessary for solving all the puzzles in our dataset.

\noindent{\textbf{Shape/Size:}} Shape and size of the visual components are two aspects that are essential in several puzzles. This category includes the understanding of both absolute and relative shapes and sizes of the puzzle components. \textit{Chain Link}, \textit{Tower of Hanoi}, and \textit{Wood Slide} fall under this category. 

\noindent{\textbf{Text:}} This category includes puzzles for which some features are shown as optical characters or embedded texts in the image. In some of our puzzles, the embedded text contains important information that must be used to correctly solve the question. The \textit{Calendar}, \textit{Clock}, and \textit{Water Jugs} puzzles fall under this category. An interesting case is the \textit{Map Colour} puzzle which has some embedded text in the image, but they are just given as additional reference and are not necessary for solving the core question.

\begin{table*}[t!]
\small
\centering
\resizebox{\textwidth}{!}{
\begin{tabular}{l|cccc|ccccccc}
\toprule[1pt]
\multirow{2}{*}{\textbf{Puzzle}} & \multicolumn{4}{c|}{\textbf{Visual Features}} & \multicolumn{7}{c}{\textbf{Algorithmic Features}}\\
& Colour & Position & Shape/Size & Text & Arithmetic & Boolean Logic & Combinatorics & Graphs  & Optimization & Search & Sets \\
\midrule 
Board Tiling     & \greencheck & \greencheck & \redcross   & \redcross   & \greencheck & \greencheck & \redcross   & \redcross   & \redcross   & \redcross   & \redcross   \\
Calendar         & \redcross   & \greencheck & \redcross   & \greencheck & \greencheck & \redcross   & \redcross   & \redcross   & \redcross   & \redcross   & \redcross   \\
Chain Link       & \redcross   & \greencheck   & \greencheck & \redcross   & \greencheck & \redcross   & \redcross   & \redcross   & \greencheck & \redcross   & \greencheck \\
Checker Move     & \greencheck & \greencheck & \redcross   & \redcross   & \greencheck & \greencheck & \redcross   & \redcross   & \redcross   & \greencheck & \redcross   \\
\greyrule
Clock            & \redcross   & \greencheck & \greencheck & \greencheck & \greencheck & \redcross   & \redcross   & \redcross   & \redcross   & \redcross   & \redcross   \\
Colour Hue       & \greencheck & \greencheck & \redcross   & \redcross   & \greencheck & \redcross   & \redcross   & \redcross   & \greencheck & \redcross   & \redcross   \\
Map Colour       & \greencheck & \greencheck & \greencheck & \greencheck & \greencheck & \greencheck & \greencheck & \greencheck & \redcross   & \greencheck & \redcross   \\
Maze Solve       & \greencheck & \greencheck & \redcross   & \redcross   & \greencheck & \redcross   & \redcross   & \greencheck & \greencheck & \greencheck & \redcross   \\
\greyrule
Move Box         & \greencheck & \greencheck & \redcross   & \redcross   & \greencheck & \redcross   & \redcross   & \greencheck & \greencheck & \greencheck & \redcross   \\
N-Queens         & \redcross   & \greencheck & \redcross   & \redcross   & \greencheck & \greencheck & \redcross   & \redcross   & \redcross   & \greencheck & \redcross   \\
Number Slide     & \redcross   & \greencheck & \redcross   & \greencheck & \greencheck & \redcross   & \greencheck & \redcross & \greencheck & \greencheck & \greencheck \\
Rotten Fruits    & \redcross   & \greencheck & \redcross   & \redcross   & \greencheck & \greencheck & \redcross   & \greencheck  & \greencheck & \greencheck & \redcross   \\
\greyrule
Rubik's Cube     & \greencheck & \greencheck  & \redcross  & \redcross   & \greencheck & \redcross   & \redcross   & \greencheck & \redcross   & \redcross   & \redcross   \\
Think A Dot      & \greencheck & \greencheck & \greencheck & \redcross   & \greencheck & \greencheck & \redcross   & \redcross   & \redcross   & \redcross   & \greencheck \\
Tower of Hanoi   & \redcross   & \greencheck & \greencheck & \redcross   & \greencheck & \greencheck & \redcross   & \redcross   & \greencheck & \greencheck & \redcross   \\
\greyrule
Water Jugs       & \redcross   & \greencheck   & \greencheck & \greencheck & \greencheck & \greencheck & \redcross   & \redcross   & \greencheck & \greencheck & \redcross   \\
Wheel of Fortune & \redcross   & \greencheck & \greencheck & \greencheck & \greencheck & \redcross   & \redcross   & \redcross   & \redcross   & \redcross   & \redcross   \\
Wood Slide       & \redcross   & \greencheck & \greencheck & \redcross   & \greencheck & \redcross   & \redcross   & \redcross & \greencheck & \greencheck & \redcross   \\
\bottomrule
\end{tabular}
}
\caption{Ontological categorization of the puzzles. We divide the table in groups for ease of reading.}
\label{tab:ontology}
\end{table*}

\subsection{Algorithmic Features}
We now detail the algorithmic features present in our puzzles. 
The ontological categorization presented below is strictly limited to obtaining the solutions for the puzzles i.e. for answering the questions for the puzzle instances. The algorithmic categories are not mutually exclusive, as we need to use two or more categories to derive the answer for most puzzles. We show examples of all the algorithmic categories in \Cref{fig:teaser}. The categorization is as follows:

\noindent{\textbf{Arithmetic:}}
This category includes puzzles where the application of basic mathematical operations such as addition, multiplication, counting, modular arithmetic, etc. are required. These are fundamental mathematical operations that are useful in all the puzzles. For example, addition and subtraction operations are required in the \textit{Water Jugs} puzzle, 
and a modular arithmetic strategy is required in the \textit{Clock} and \textit{Calendar}. 

\noindent{\textbf{Boolean Logic:}} The puzzles for which application of some form of boolean logic is necessary for answering the question. The logical statement may take various forms, for instance: checking if the number of occurrences of two colours is equal or not in \textit{Board Tiling}, flipping a disc colour whenever a ball passes through it in \textit{Think A Dot}. 

\noindent{\textbf{Combinatorics:}}
This category includes puzzles that deal with counting combinations and permutations of its states. The general form of the question is about how many unique configurations can be reached after performing a sequence of operations. 
These kinds of problems are generally easier to solve with computer algorithms when there are a lot of constraints on how the configurations can be created. Our \textit{Map Colour} and \textit{Number Slide} puzzles come under this category. 

\noindent{\textbf{Graphs:}}
This category includes puzzles that can be represented as a graph data structure on which specific graph algorithms can be applied to solve the problem. Such algorithms include graph colouring, graph traversal, etc. This category includes puzzles such as \textit{Map Colour}, \textit{Maze Solve}, and \textit{Rotten Fruits}. 

\noindent{\textbf{Optimization:}} 
Optimization is a crucial area of mathematics and computer science that mainly deals with finding optimal solutions to a problem. Our puzzles involve problems such as solving a task in the minimum possible time, reaching a goal from an initial state in minimum steps, optimal sorting, summation maximization, etc. This category includes puzzles such as \textit{Chain Link}, \textit{Colour Hue}, \textit{Tower of Hanoi}, and \textit{Water Jugs}.

\noindent{\textbf{Search:}} 
Search algorithms are generally used to retrieve information stored within a particular data structure, or to calculate the search space of a particular problem domain. Although there are many different kinds of search algorithms, our puzzles are constrained to breadth-first search and exhaustive search. Examples of such puzzles include \textit{Checker Move}, \textit{Move Box}, \textit{Wood Slide}, etc. 

\noindent{\textbf{Sets:}}
Many problems in computer science deal with sets of objects where you have to consider the identical nature of some objects and the equivalence of some positions or configurations. We include some puzzles in our dataset where such properties can be or have to be used to solve the problem. This category includes the \textit{Checker Move}, \textit{Number Slide}, and \textit{Think A Dot} puzzles. 

\section{From the Ontology to Creating \dataset{}}
\label{sec:puzzles}
We create a total of 18 different puzzles for our dataset following various feature combinations from our ontology (\Cref{tab:ontology}). Many of these puzzles are popular in various recreational or academic settings. 

\paragraph{Generating puzzle images and curating solutions.}
Instances for each puzzle are created automatically from the human-written code of the particular puzzle which includes: 
\begin{enumerate}
    \item Code to generate the image file that we use as visual context. We use the \code{matplotlib} library in \code{Python} to generate the images.
    \item Code for applying the correct algorithm for finding the solution. The algorithmic solutions to these puzzles are well-known and are easy to verify against literature and other popular coding resources such as LeetCode\footnote{\url{https://leetcode.com/}}.
\end{enumerate}

\paragraph{Creating language context.} Finally, we also carefully curate the language context (in English) for each puzzle using puzzle-specific templates. We make sure that the template describes all the necessary information and rules required for solving the puzzle. The templates were written by two authors of the paper and were verified by the other authors.

\paragraph{Extensibility.}
As mentioned earlier, the number of instances and the difficulty level of the puzzles can be scaled arbitrarily to create a large and challenging dataset. For now, we have created 100 instances for each puzzle. These instances are analogous to different \textit{test cases} of the puzzle, i.e. they have different input combinations, initial and goal states, etc. Reliably solving all the instances would require finding the exact algorithm to use and then applying it accurately. This is akin to how we verify the accuracy of a computer program aiming to solve a particular task through a broad range of test cases.

\paragraph{Difficulty.} Our current version of \dataset{} contains problems with easy to moderate difficulty, many of which can be solved by humans in hand without applying the underlying algorithm. However, if the problems are made harder then the algorithmic solution would be easier to find.

\paragraph{Dataset Size.}
In total, we have 1800 instances from the 18 different puzzles. We consider the full dataset as an evaluation-only benchmark. We describe the details and the creation process of some of the puzzles next. A detailed description of all the puzzles and their creation process can be found in the \Cref{sec:appendix:dataset}.
\subsection{Some Example Puzzles in Our Dataset}
\paragraph{Board Tiling.}\label{sec:board_tile}
\begin{figure}[t]
  \centering
  \includegraphics[width=0.3\textwidth]{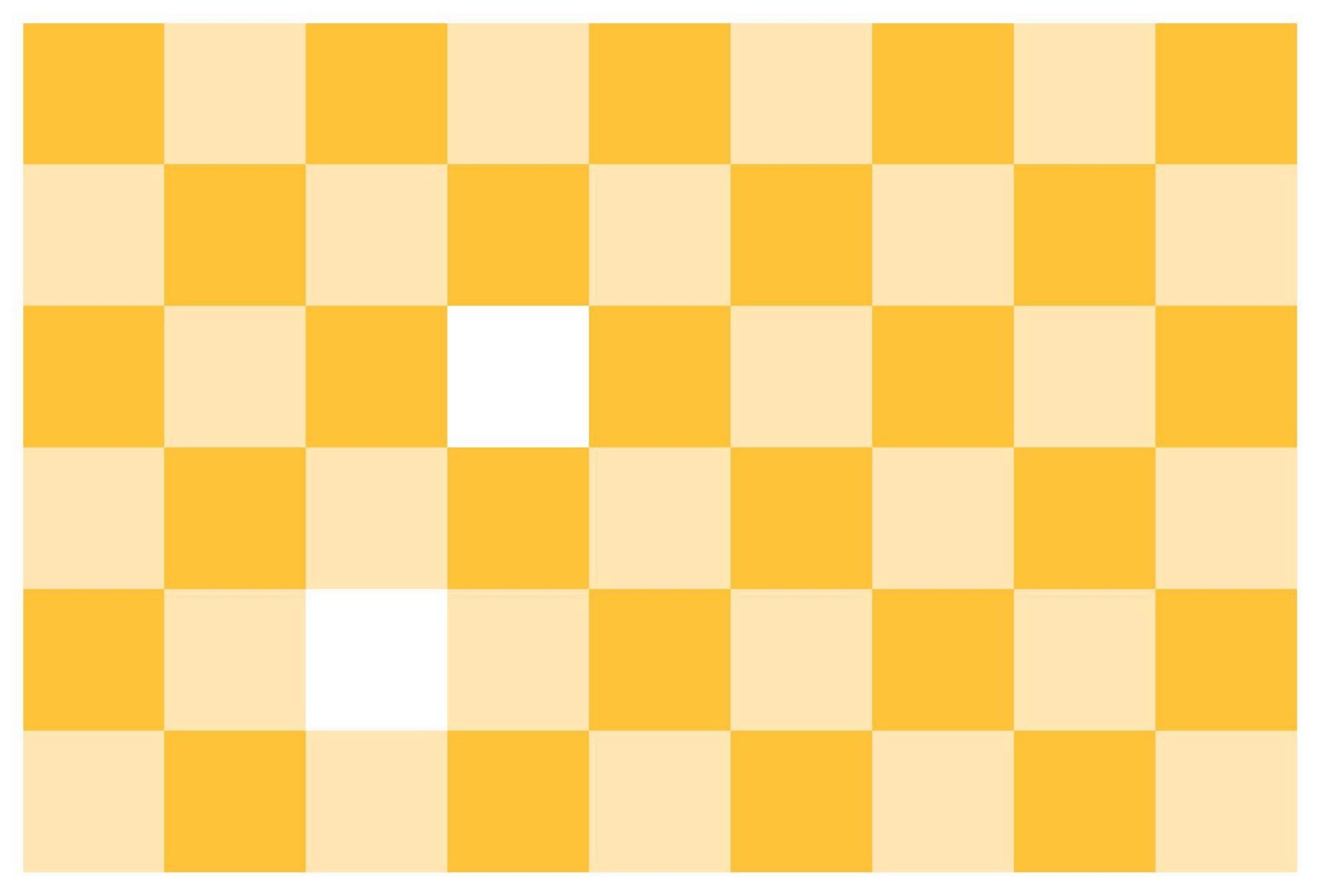}
  \caption{\footnotesize{\code{\textbf{Question:} The checkerboard shown in the image was originally of 6 * 9 in dimension having a total of 54 squares. It uses two colours of squares, one light yellow and one dark yellow, in a chequered pattern. Two of the squares have been removed from the board in the position of the white coloured cells, as shown in the image. You have 26 dominoes of size 2 * 1. You can use them as is or you can rotate them to use as a 1 * 2 domino. Is it possible to place all the 26 dominoes in the checkerboard to exactly cover all the remaining 52 squares? Answer Yes or No. \textbf{Gold Answer:} Yes}}}
  \label{fig:board_tile}
\end{figure}

The puzzle is inspired by the Mutilated chessboard problem, originally posed by Max Black~\citep{black1948critical}. The puzzle belongs to the class of domino tiling problems.
A general result from ~\citet{mendelsohn2004tiling} states that: a checkerboard originally had an even (odd) number of squares. Two (one) randomly chosen squares are now removed from the board. If the mutilated board has $2 * m$ squares then it is tileable with $m$ dominoes of size $2 * 1$ if and only if it has an equal number of dark and light squares.
We choose the number of rows and columns in our checkerboard randomly between $4$ to $9$. We randomly remove $1$ or $2$ squares if the board has an odd or even number of squares, respectively. 
We determine the yes or no tilebality answer based on the general result. We show an example of the puzzle in \Cref{fig:board_tile}.

\paragraph{Colour Hue.}

\begin{figure}[t]
  \centering
  \includegraphics[width=0.35\textwidth]{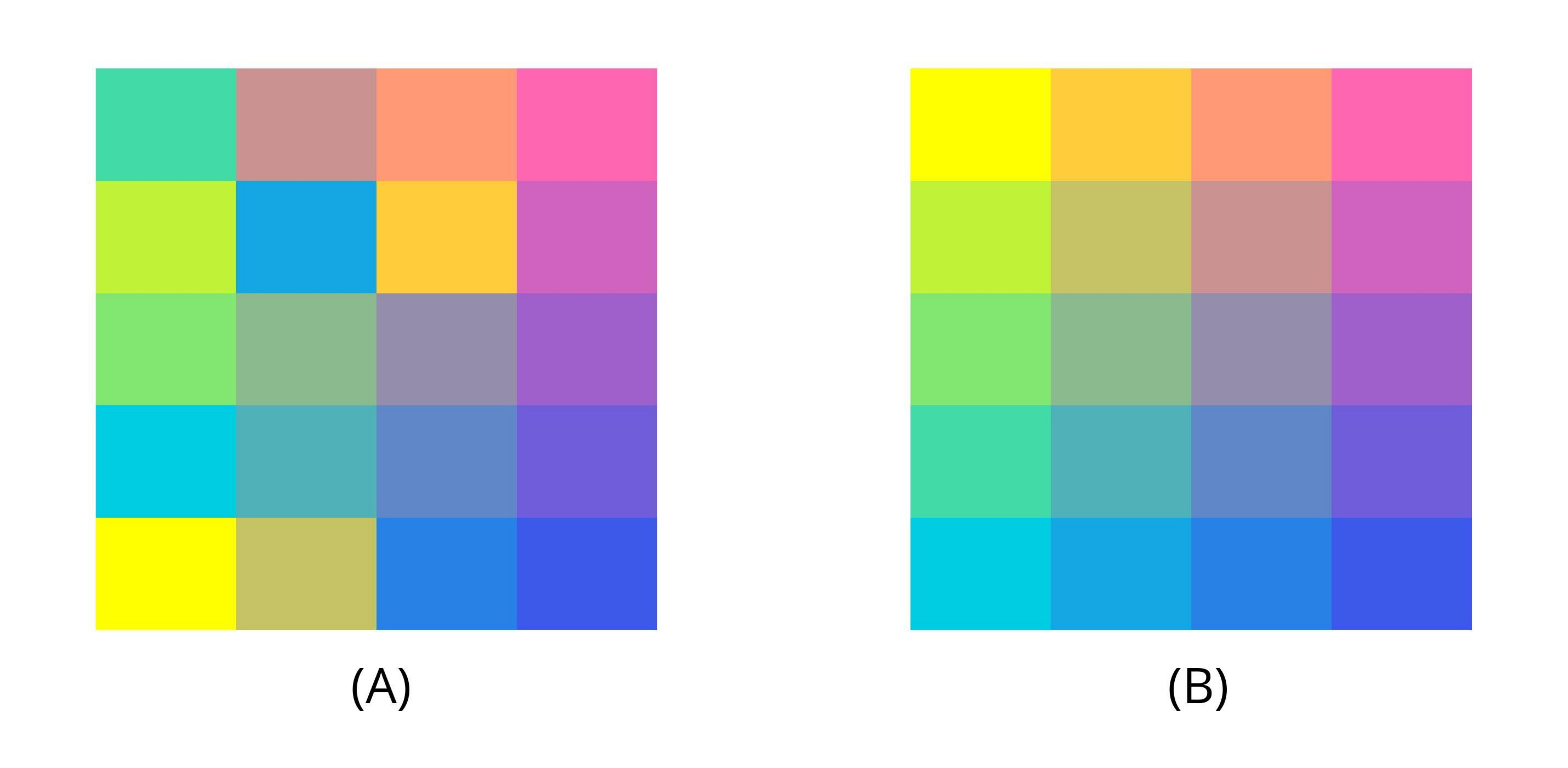}
  \caption{\footnotesize{\code{\textbf{Question:} A 5 * 4 board consists of 20 different coloured tiles. A random state of the board is shown in (A). The ideal state of the board is shown in (B). A swap consists of selecting any two tiles in the board and switching their positions. What is the minimum number of swaps required to restore the ideal state of the board from (A)? \textbf{Gold Answer:} 4}}}
  \label{fig:colour_hue}
\end{figure}

A rectangular board contains of $m * n$ coloured tiles arranged in a rectangular grid. The board has an ideal state. 
A non-ideal state of the board is created by shuffling some tiles. 
We ask how many minimum tile swaps are required to reach the ideal state from the non-ideal state. The answer can be determined using the selection sort algorithm, which minimizes the number of swaps required to sort an unsorted array. 
We considered the non-ideal state as the unsorted state and the ideal state as the sorted state to find the answer.
We show an example in \Cref{fig:colour_hue}.

\paragraph{Map Colouring.}

\begin{figure}[h!]
  \centering
  \includegraphics[width=0.35\textwidth,height=4.5cm]{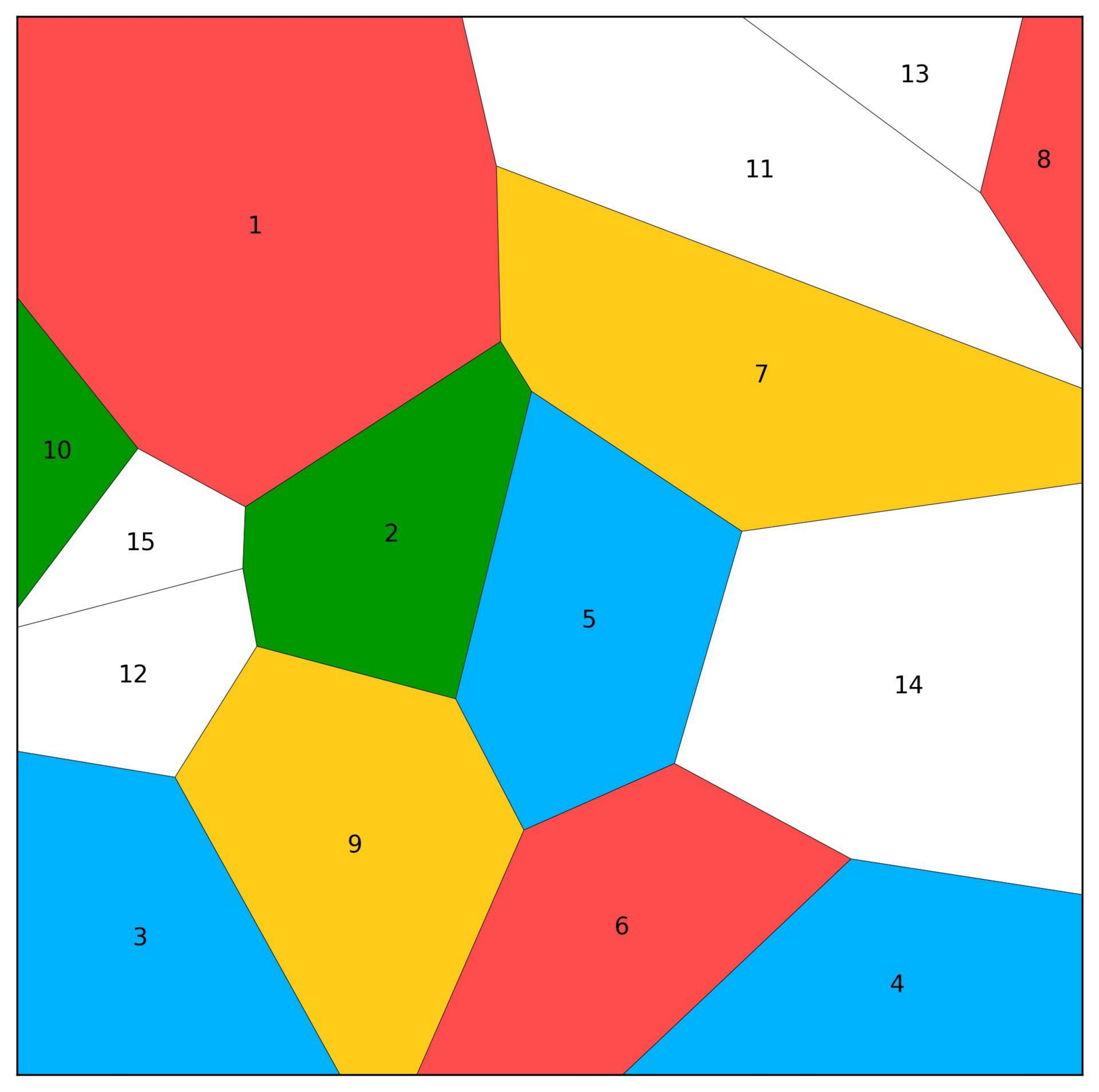}
  \caption{\footnotesize{\code{\textbf{Question:} You are given an incomplete map of a country having 15 different regions. The objective is to colour the regions of the map using only the four available colours: red, green, blue and yellow, such that no two adjacent regions have the same colour. Adjacent regions are defined as two regions that share a common boundary of non-zero length. The regions indicated by numbers 1 to 10 have already been coloured, as shown in the image. The regions indicated by numbers 11 to 15 are shown in white as they are yet to be coloured. You need to assign colours to these regions in a way such that it doesn't violate the objective. Each unique colour combination of the regions would result in a unique complete map. How many unique complete maps can be created by colouring all the white regions starting from the given incomplete map? \textbf{Gold Answer:} 8}}}
  \label{fig:map}
\end{figure}

The four colour theorem is a famous result in mathematics that states that four colours are sufficient to colour the regions of any planar map such that no two adjacent regions have the same colour. 
We create a Voronoi diagram 
~\citep{Voronoi1908} followed by polygon clipping ~\citep{sutherland1974reentrant} 
to create the regions of the map.
We use a graph colouring strategy based on Algorithm X \citep{knuth2000dancing} with four colours to find the exhaustive solutions for colouring the map. We fix the colours of some of the regions of the map and find out the number of ways the remaining regions can be coloured from the exhaustive list of solutions. We show an example of this puzzle in \Cref{fig:map}.

\begin{table*}[ht]
\small
\centering
\resizebox{0.9\textwidth}{!}{
\begin{tabular}{l|ccc|ccc|ccc|c}
\toprule[1pt]
\multirow{2}{*}{\textbf{Puzzle}} & \multicolumn{3}{c|}{\textbf{CoT}} & \multicolumn{3}{c|}{\textbf{eCoT}} & \multicolumn{2}{c}{\textbf{I-BLIP}} & \textbf{LLaVA} & \multirow{2}{*}{\textbf{Average}} \\
& \textbf{GPT-4V} & \textbf{Gemini Pro} & \textbf{Claude 3} & \textbf{GPT-4V} & \textbf{Gemini Pro} & \textbf{Claude 3} & 7B & 13B & 13B & \\
\midrule 
Board Tiling & 48 & 43 & 48 & 45 & 50 & 47 & 52 & 52 & 54 & 48.8 \\
Calendar & 57 & 33 & 39 & 51 & 30 & 42 & 18 & 21 & 31 & 35.8 \\
Chain Link & 21 & 30 & 23 & 31 & 28 & 29 & 29 & 24 & 31 & 27.3 \\
Checker Move & 33 & 27 & 35 & 34 & 25 & 33 & 34 & 15 & 27 & 29.2\\
\greyrule
Clock & 25 & 32 & 37 & 29 & 30 & 34 & 28 & 26 & 10 & 27.9 \\
Colour Hue & 27 & 29 & 16 & 23 & 21 & 31 & 21 & 18 & 22 & 23.1 \\
Map Colour & 34 & 28 & 33 & 34 & 38 & 29 & 17 & 25 & 29 & 29.7 \\
Maze Solve & 30 & 38 & 23 & 31 & 32 & 25 & 27 & 21 & 27 & 28.2 \\
\greyrule
Move Box & 29 & 32 & 19 & 25 & 26 & 14 & 24 & 28 & 20 & 24.1 \\
N-Queens & 14 & 35 & 15 & 24 & 23 & 13 & 15 & 10 & 27 & 19.6 \\
Number Slide & 41 & 38 & 43 & 32 & 31 & 46 & 27 & 35 & 32 & 36.1 \\
Rotten Fruits & 27 & 29 & 40 & 42 & 25 & 39 & 27 & 33 & 29 & 32.3 \\
\greyrule
Rubik's Cube & 48 & 34 & 39 & 43 & 26 & 36 & 41 & 41 & 37 & 38.3 \\
Think A Dot & 31 & 31 & 53 & 40 & 33 & 47 & 34 & 34 & 41 & 38.2 \\
Tower of Hanoi & 17 & 29 & 22 & 21 & 29 & 23 & 29 & 22 & 26 & 24.2 \\
\greyrule
Water Jugs & 17 & 13 & 37 & 17 & 15 & 28 & 41 & 42 & 21 & 25.7 \\
Wheel of Fortune & 25 & 27 & 19 & 28 & 24 & 20 & 23 & 19 & 27 & 23.6 \\
Wood Slide & 21 & 15 & 16 & 21 & 20 & 26 & 37 & 30 & 22 & 23.1 \\
\midrule
\textbf{Average} & 30.3 & 30.2 & 30.9 & \textbf{31.7} & 28.1 & 31.2 & 29.1 & 27.6 & 28.5 & - \\
\bottomrule
\end{tabular}
}
\caption{\footnotesize{Accuracy scores across all the puzzles for the various multimodal language models. We divide the table into groups for ease of reading. Claude 3 indicates the Claude 3 Opus model and I-BLIP indicates the Instruct-BLIP model.}}
\label{tab:results}
\end{table*}

\section{Experiments on \dataset{}}
\subsection{Setup and Baselines}
We perform all our experiments on a multi-choice question-answering setup. We create three negative answer choices for each instance 
by using heuristics such as randomly sampling numbers within the same magnitude as the gold answer. More details can be found in \Cref{sec:appendix:mcqa}.
In total, we thus have four answer choices - one gold positive and three random negatives, for all puzzles except one. The exception is the \textit{Board Tiling} puzzle, where we have \textit{Yes} and \textit{No} as the possible choices.
We evaluate various closed and open-source multimodal language models on our dataset. We consider the following models: GPT-4V~\cite{gpt4vision}, Gemini Pro~\cite{geminipro}, Claude 3 Opus\footnote{\url{https://www.anthropic.com/news/claude-3-family}} as the closed models; InstructBLIP Vicuna 7B and 13B~\cite{Dai2023InstructBLIPTG}, and LLaVA-1.5 13B~\cite{liu2023improved} as the open-sourced models. We use the accuracy of predicting the final answer as the evaluation metric.



\subsection{Prompting Strategy}
\paragraph{GPT4V, Gemini, Claude, and LLaVA:} We use the zero-shot chain-of-thought (CoT) technique for these models. The objective is to generate the reasoning steps and then the final answer from the image, question and multiple answer choices. We perform experiments with two types of CoT settings using the following prompting instructions: (i) \code{Let's think step by step} \citep{NEURIPS2022_8bb0d291}, and (ii) \code{Let's describe the image first and think step by step}. We use the notation CoT to describe the first setting and Elaborate CoT or eCoT to describe the second setting. For LLaVA, we only use the CoT setting.
We concatenate the question, answer choices and the prompting instruction to create the final prompt. An example prompt from the CoT setup is as follows: \code{Question: You are playing a Tower of Hanoi game $\dots$ Options: (A) 1 (B) 2 (C) 6 (D) 3. Answer: Let's think step by step.} We create the prompt for eCoT in a similar fashion with its respective prompting instruction. 
We generate the output using a temperature of 0 which is equivalent to greedy decoding 
\citep{NEURIPS2022_8bb0d291,wei2022chain}.

\paragraph{Instruct-BLIP:} We follow the multi-choice question-answering setup recommended in the original Instruct-BLIP work ~\cite{Dai2023InstructBLIPTG} for evaluation. The prompt is: \code{Question: You are playing a Tower of Hanoi $\dots$ Options: (a) 1 (b) 2 (c) 6 (d) 3. Short Answer:} The output generated from the model is constrained to be within the answer choices using a vocabulary restriction method. The answer choice with the highest log-likelihood is chosen as the prediction.

\subsection{Main Results}
\begin{table*}[ht]
\centering
\resizebox{0.9\textwidth}{!}{
\begin{tabular}{l|ccc|ccc|cccc}
\toprule[1pt]
& \multicolumn{3}{c|}{\textbf{CoT}} & \multicolumn{3}{c|}{\textbf{eCoT}} & \multirow{2}{*}{\textbf{I-BLIP 7B}} & \multirow{2}{*}{\textbf{I-BLIP 13B}} & \multirow{2}{*}{\textbf{LLaVA 13B}}\\
& \textbf{GPT-4V} & \textbf{Gemini Pro} & \textbf{Claude 3} & \textbf{GPT-4V} & \textbf{Gemini Pro} & \textbf{Claude 3}\\
\midrule
\textbf{Visual Features} &  &  &  &  &  & & & & \\
\quad Colour & \textbf{35.0} & 32.8 & 33.2 & 34.4 & 31.4 & 32.8 & 31.2 & 29.2 & 32.1 \\
\quad Position & 30.3 & 30.2 & 30.9 & \textbf{31.7} & 28.1 & 31.2 & 29.1 & 27.6 & 28.5 \\
\quad Shape/Size & 23.9 & 25.6 & \textbf{30.0} & 27.6 & 27.1 & 29.5 & 29.8 & 27.8 & 25.9 \\
\quad Text & 33.2 & 28.5 & \textbf{34.7} & 31.8 & 28.0 & 33.2 & 25.7 & 28.0 & 25.0  \\
\greyrule
\textbf{Algorithmic Features} &  &  &  &  &  & &  \\
\quad Arithmetic & 30.3 & 30.2 & 30.9 & \textbf{31.7} & 28.1 & 31.2 & 29.1 & 27.6 & 28.5 \\
\quad Boolean Logic & 27.6 & 29.4 & \textbf{35.4} & 32.1 & 29.8 & 32.4 & 31.1 & 29.1 & 31.8 \\
\quad Combinatorics & 37.5 & 33.0 & \textbf{38.0} & 33.0 & 34.5 & 37.5 & 22.0 & 30.0 & 30.5 \\
\quad Graphs & 33.6 & 32.2 & 30.8 & \textbf{35.0} & 29.4 & 28.6 & 27.2 & 29.6 & 28.4 \\
\quad Optimization & 25.6 & 28.1 & 26.6 & 27.0 & 25.2 & 29.0 & \textbf{29.1} & 28.1 & 25.6 \\
\quad Search & 26.3 & \textbf{28.4} & 28.3 & 28.1 & 26.4 & 27.6 & 27.8 & 26.1 & 26.0 \\
\quad Sets & 31.0 & 33.0 & 39.7 & 34.3 & 30.7 & \textbf{40.7} & 30.0 & 31.0 & 34.7 \\
\bottomrule
\end{tabular}
}
\caption{\footnotesize{Accuracy scores across all the ontological categories for the various multimodal language models.}}
\label{tab:results:ontology}
\end{table*}

We report the main results for all the models across all the puzzles in \Cref{tab:results}. The Board Tiling puzzle has a random baseline performance of 50\%. All other puzzles have a random baseline performance of 25\%. The overall random baseline stands at 26.4\%. We notice that the performance in a significant number of these puzzles across all the models is close to the random baseline of 50\% and 25\%. In CoT setup, Claude 3 Opus acheives the best average score of 30.9\%.
The GPT-4V model in the eCoT setup achieves the best score overall with an average accuracy of 31.7\%, which is only around 5\% better than random. The other models perform slightly poorer, with Gemini Pro obtaining a best of 30.2\%, Claude 3 obtaining a best of 31.2\%, Instruct-BLIP obtaining a best of 29.1\% with the 7B model, and LLaVA obtaining 29.1\%. 

We report the average score of the models for each puzzle in the right-most column of \Cref{tab:results}. This helps us find which puzzles are easier and which are difficult for models. We found that \textit{Rubik's Cube} and \textit{Think A Dot}
have the highest average score over random, implying that these two puzzles are found by models to be comparatively the easiest. Conversely, the average performance in \textit{N-Queens} and \textit{Colour Hue} 
are the lowest, signifying that models found them to be the hardest.

In terms of absolute score, we find the highest performing experiment to be GPT-4V CoT in the \textit{Calendar} puzzle, where it achieves an accuracy of 57\%. 
We do not find any puzzle where the accuracy is higher than 60\%. We conclude that large multimodal language models find the task of visual algorithmic problem-solving to be very challenging. Even though they have achieved remarkable performance in many tasks, they still have some way to go in performing complex reasoning tasks defined over vision, language, mathematics, and algorithms.

\subsection{Ontological Analysis}
We present the results across ontological categories in \Cref{tab:results:ontology}. The obtained results suggest some interesting patterns across the ontological structure. Closed models mostly perform better across the visual features of colour, position, and text but perform poorer on shape/size. 

The best-performing model GPT-4V eCoT performs much higher in algorithmic topics such as combinatorgraphs and sets compared to topics such as optimization and search. Results suggest that the optimization and search topics are the most difficult topics in general across all the models.

\subsection{Reasoning with Guided Vision}
Our current experimental setup doesn't disentangle the visual perception stage and algorithmic reasoning stage. In the original setup, models must identify the various aspects and characteristics of the visual context before the appropriate algorithm can be applied. To minimize the effect of the bottleneck in the visual perception stage, we conduct a guided vision experiment, where we additionally provide detailed descriptions of the image as part of the language context. In this setup, errors from the visual perception stage are minimized, so models can only focus on the algorithmic stage for solving the question.  

We report the results in \Cref{tab:results:guided_vision}. The upper part of the table constitutes the puzzles where the algorithmic reasoning is difficult, as even with language-guided visual context, the model cannot improve its scores. For GPT-4V models, the lower part of the table indicates puzzles where the guided vision setup helps in a large improvement of performance, suggesting there is a significant bottleneck in the visual perception stage. However, even then the numbers don't go anywhere close to 100, suggesting that the algorithmic reasoning stage presents substantial challenges even in the presence of gold context.

\begin{table}[t!]
\small
\centering
\resizebox{0.4\textwidth}{!}{
\begin{tabular}{l|cc|cc}
\toprule[1pt]
& \multicolumn{2}{c|}{\textbf{GPT-4V}} & \multicolumn{2}{c}{\textbf{Gemini Pro}} \\
& w/o & w/ & w/o & w/ \\
\midrule 
Calendar & 51 & 43 & 30 & 31 \\
Water Jugs  & 17 & 23 & 15 & 12 \\
\greyrule
Checker Move  & 34 & \textbf{45} & 25 & \textbf{32} \\
Clock  & 29 & \textbf{74} & 21 & \textbf{43} \\
Move Box  & 25 & \textbf{61} & 26 & 21 \\
Number Slide  & 32 & \textbf{53} & 31 & \textbf{39} \\
Tower of Hanoi & 19 & \textbf{32} & 18 & 19 \\
\bottomrule
\end{tabular}
}
\caption{\footnotesize{Reasoning with Guided Vision. w/o and w/ indicate the without and with the guided vision context.}}
\label{tab:results:guided_vision}
\end{table}


\section{Conclusion}
In this study, we introduce \dataset{}, a novel dataset comprising puzzles that demand multimodal reasoning over vision, language and algorithms. Unlike existing multimodal reasoning datasets such as ScienceQA or MMMU, the challenges in \dataset{} do not hinge on possessing domain-specific knowledge for reasoning. Instead, solving the problems in this dataset necessitates applying logical algorithmic steps to novel problem scenarios. To construct this dataset, we devised an ontology encompassing i) visual reasoning features such as colours, positions, shapes and sizes of puzzle components; and ii) algorithmic reasoning features such as arithmetic, combinatorics, optimization, etc. Each puzzle in our dataset comprises multiple categories from the visual and algorithmic ontology. Through rigorous experiments with LLMs such as GPT-4V, Gemini-Pro, Claude 3, InstructBlip, and LLaVA, we observed a consistent struggle in achieving satisfactory performance on \dataset{}, emphasizing the formidable hurdles in multimodal reasoning for algorithmic puzzle-solving. Given that each puzzle is annotated with specific visual and algorithmic features based on our ontology, we draw insights into the precise skills lacking in LLMs to address particular problems. We also analyzed whether the models falter on this dataset due to deficiencies in the visual recognition setup or the algorithmic reasoning setup. While furnishing the models with accurate visual guidance of the puzzles enhances performance in some cases, achieving high-level performance in our proposed complex reasoning tasks remains an elusive goal.

\section{Limitations}
We have currently considered many puzzles that are popular in various recreational or academic settings. There are also other interesting puzzles beyond our compiled list which can be used to assess the complex reasoning abilities of LLMs. We aim to explore them and enlarge our suite of puzzles as future work. Additionally, the construction of the algorithmic ontology could be expanded to consider more fine-grained categories. 

We have also prompted the language models to generate the reasoning steps and the answer through natural language. The class of models that can also generate code could also be explored in the future for solving our proposed visual algorithmic reasoning tasks. Such models might be able to generate the algorithmic steps required for solving the problem through code. We aim to explore such models as part of future work. 


\bibliography{custom}

\appendix
\section{Puzzles}
\label{sec:appendix:dataset}

We consider the following 18 puzzles:

\subsection{Board Tiling}\label{sec:board_tile}
\begin{figure}[t]
  \centering
  \includegraphics[width=0.4\textwidth]{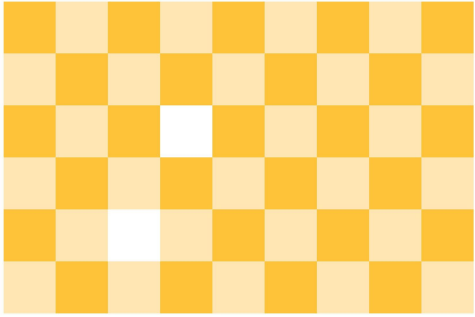}
  \caption{\footnotesize{\code{\textbf{Question:} The checkerboard shown in the image was originally of 6 * 9 in dimension having a total of 54 squares. It uses two colours of squares, one light yellow and one dark yellow, in a chequered pattern. Two of the squares have been removed from the board in the position of the white coloured cells, as shown in the image. You have 26 dominoes of size 2 * 1. You can use them as is or you can rotate them to use as a 1 * 2 domino. Is it possible to place all the 26 dominoes in the checkerboard to exactly cover all the remaining 52 squares? Answer Yes or No. \textbf{Gold Answer:} Yes}}}
  \label{fig:appendix:board_tile}
\end{figure}

The puzzle is inspired by the Mutilated chessboard problem, originally posed by Max Black~\citep{black1948critical}:
\textit{Suppose a standard $8 * 8$ chessboard has two diagonally opposite corners removed, leaving 62 squares. Is it possible to place 31 dominoes of size $2 * 1$ to cover all of these squares?}
The diagonally opposite corners in a standard $8 * 8$ chessboard are always of the same colour. Hence, the mutilated chessboard with $62$ squares has $30$ squares of one colour and $32$ squares of the other colour. Now, the checkered pattern of the chessboard ensures that each $2 * 1$ domino must cover $1$ dark-coloured square and $1$ light-coloured square. It is thus impossible to place the dominoes to cover all the squares since the mutilated chessboard has an unequal number of dark and light-coloured squares.

A general result from ~\citet{mendelsohn2004tiling} states that: a checkerboard originally had an even (odd) number of squares. Two (one) randomly chosen squares are now removed from the board. If the mutilated board has $2 * m$ squares then it is tileable with $m$ dominoes of size $2 * 1$ if and only if it has an equal number of dark and light squares.
We use this result to construct our puzzles. 
We choose the number of rows and columns in our checkerboard randomly between $4$ to $9$. We randomly remove $1$ or $2$ squares if the board has an odd or even number of squares, respectively. We question whether the resulting board is tileable with $m$ dominoes of size $2 * 1$. 
We determine the yes or no tilebality answer based on the general result. We show an example of the puzzle in \Cref{fig:appendix:board_tile}.

\subsection{Calendar}\label{sec:calendar}
\begin{figure}[t]
  \centering
  \includegraphics[width=0.5\textwidth]{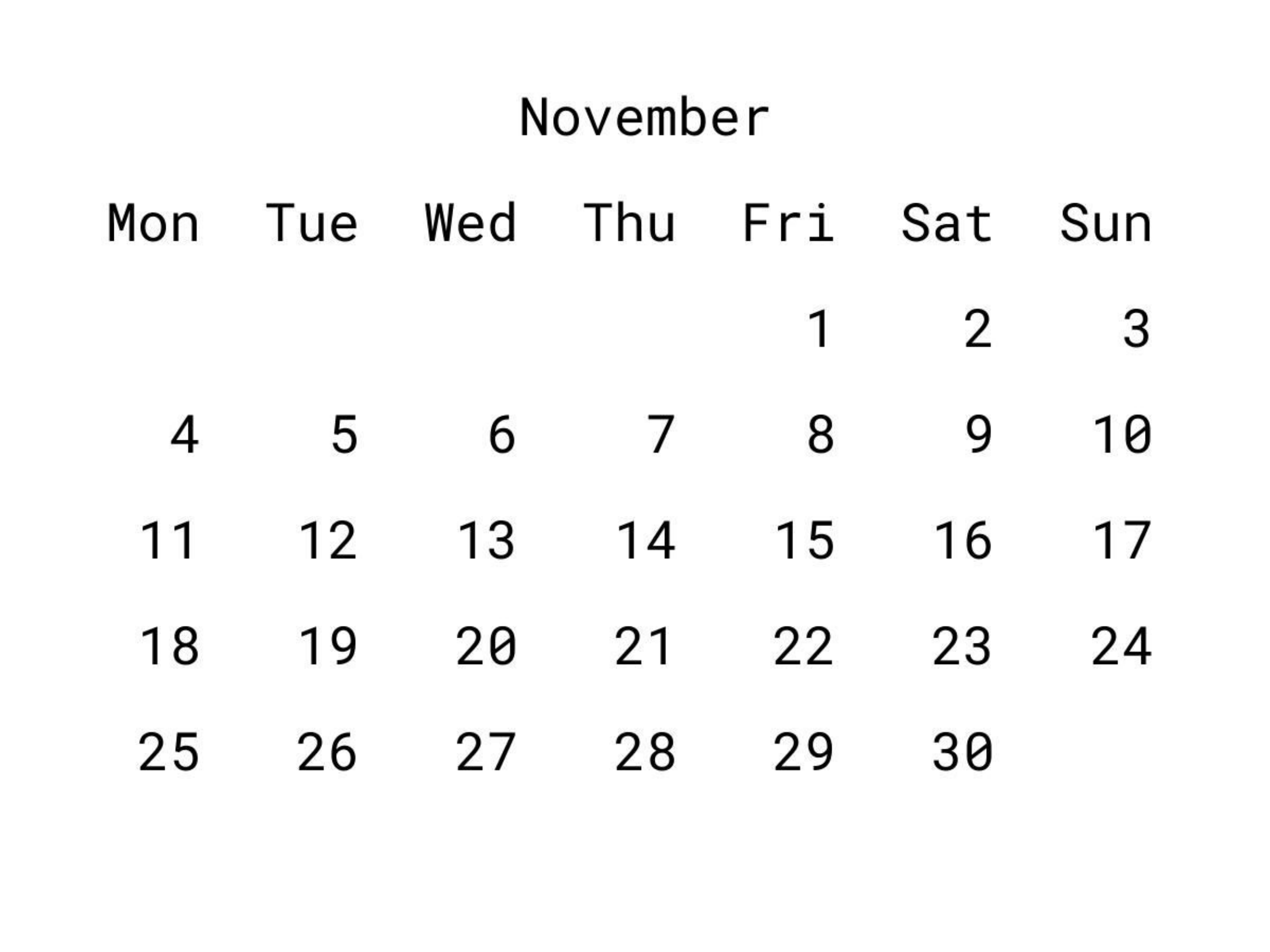}
  \caption{\footnotesize{\code{\textbf{Question:} The image shows the calendar of a month of a particular non-leap year. Which day of the week was on March 1 of that year? \textbf{Gold Answer:} Friday}}}\label{fig:appendix:calendar}
\end{figure}

We design this puzzle to evaluate the visual-temporal reasoning abilities of foundation models. We provide the calendar snapshot of a particular month as the visual context. We then ask what day of the week was a particular date in either the previous, same or the next year. We also provide information about whether the years of consideration were leap years or not to make sure that the answer is exact. We use the python \textit{calendar} module to construct the instances of the puzzle. We show an example of the puzzle in \Cref{fig:appendix:calendar}.

\subsection{Checker Move}
The puzzle generally known as \textit{Toads and Frogs}\footnote{\url{https://en.wikipedia.org/wiki/Toads_and_Frogs}} was invented by the mathematician Richard Guy. We start with a grid of length $n$, with $n-1$ checkers of either red or green colour occupying $n-1$ positions. The goal is to rearrange the checkers into a given desired position. The rearrange is constrained to some rules: 
i) green checkers only move rightward; ii) red checkers only move leftward; iii) every move is either 1) a slide to the adjacent empty square, or 2) a jump over one position to an empty square, provided the checker being jumped over is of a different colour; iv) each grid position can accommodate a maximum of one checker at any time. 

We only consider final arrangements that can be reached from the starting arrangements. We use breadth-first search constrained upon the movement rules to derive the solution.

\begin{figure}[t]
  \centering
  \includegraphics[width=0.5\textwidth]{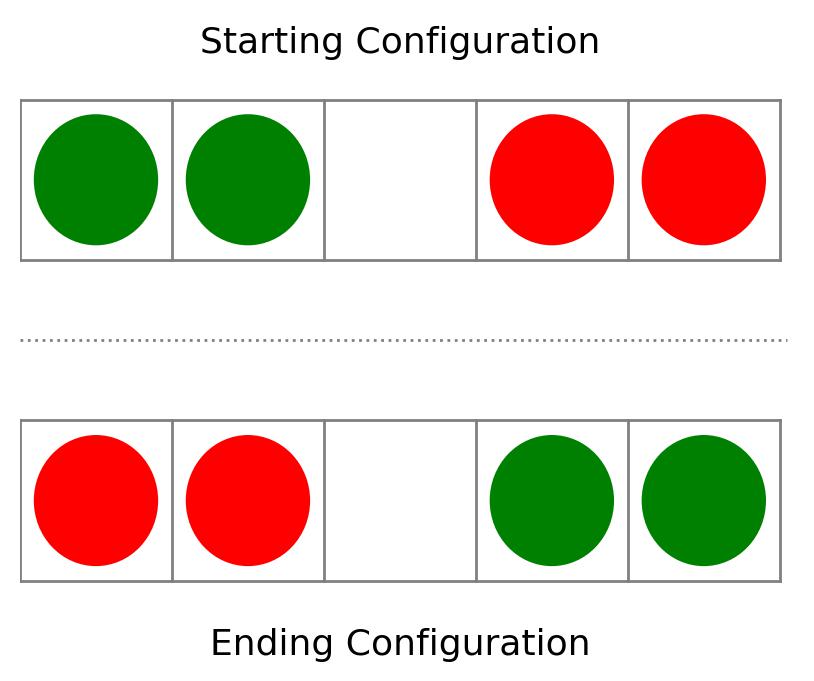}
  \caption{\footnotesize{\code{\textbf{Question:} A checker game is being played on a grid of 5 squares with 2 green and 2 red checkers. Initially, the checkers are arranged as shown in the starting configuration with the 4 checkers occupying 4 squares and one unoccupied square. Green checkers only move rightward and red checkers only move leftward. Every move is either i) a slide to the adjacent empty square, or ii) a jump over one position to an empty square, provided the checker being jumped over is of a different colour. Each square can accommodate a maximum of one checker at any time. How many moves are required to reach the ending configuration from the starting configuration following the specified rules? \textbf{Gold Answer:} 8}}}\label{fig:appendix:cheker_move}
\end{figure}

\subsection{Chain Link}

\begin{figure}[t]
  \centering
  \includegraphics[width=0.46\textwidth]{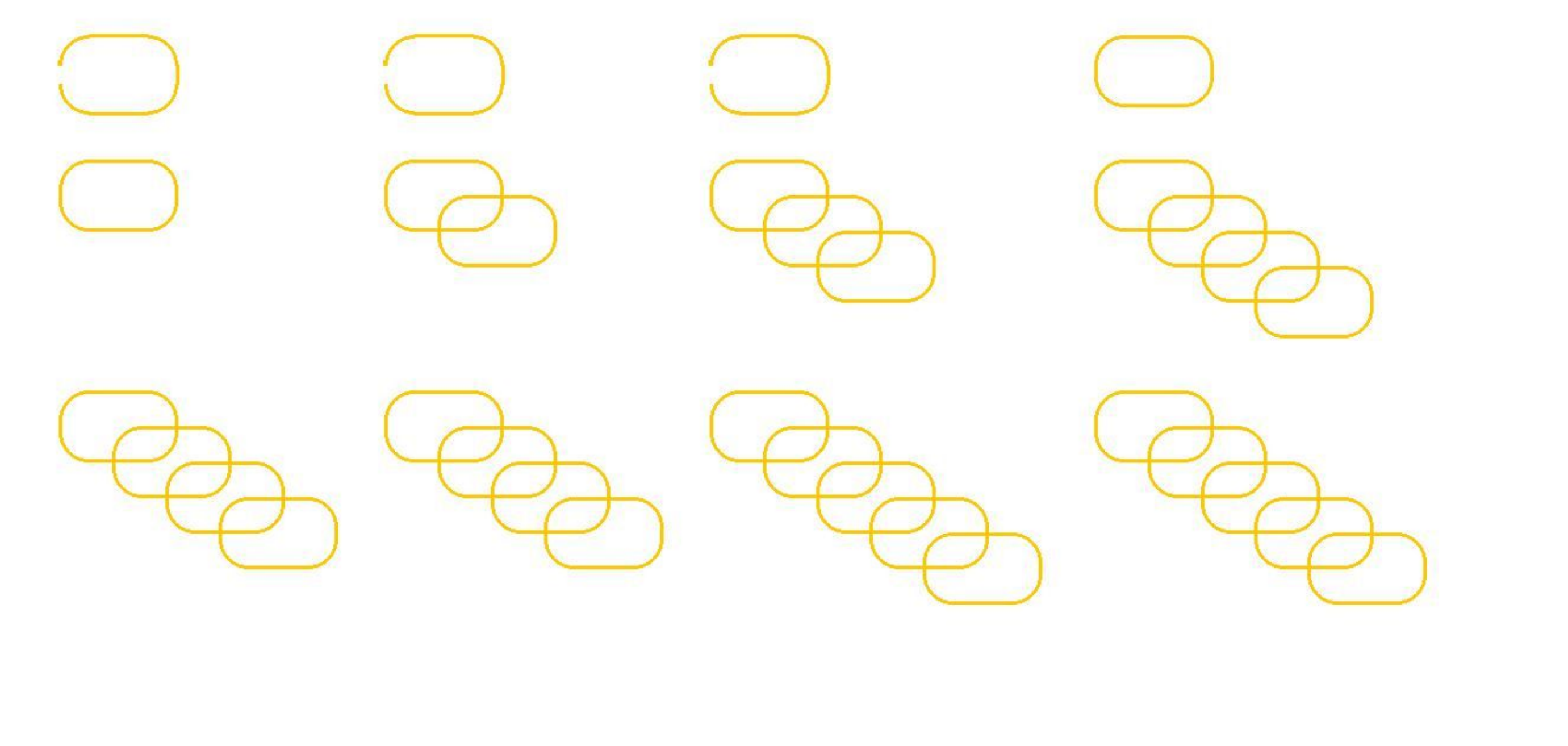}
  \caption{\footnotesize{\code{\textbf{Question:} Alice has 12 segments of chains of different lengths as shown in the image. The total length of all the segments combined is 32 pieces. She has a saw machine with which a closed piece can be cut opened. She also has a welding machine with which an open piece can be closed. Each cut takes 5 minutes and each welding takes 2 minutes. Initially, she has 3 segments each with 1 open piece as shown in the image. All the other pieces are closed. She now wants to make the longest possible necklace using all the available 32 pieces. Each piece in the necklace would be connected to exactly two other pieces. This would require cutting open some pieces and then joining all the resulting segments together. What is the minimum time in which she can create the necklace? \textbf{Gold Answer:} 34}}}
  \label{fig:appendix:chain_link}
\end{figure}

This puzzle is a modified version of a problem that appeared in ~\citet{kordemsky1992moscow}. It states that you are given chain segments of different lengths. Closed pieces can be cut open and open pieces can be welded together in a certain amount of time. You need to find out the least time required to create the longest possible circular necklace. We show an example of the puzzle in \Cref{fig:appendix:chain_link}. The puzzle can be solved optimally as follows:

\begin{enumerate}[leftmargin=4mm]
    \item Check if the number of open links is equal to or greater than the number of closed segments. Initially, in our example, the number of open links is $3$ and the number of closed segments is $9$. The condition is not satisfied and hence we move to the next step.
    \item  Cut units from the segments of the least length until the first condition is satisfied. In our example, if you cut open the two segments of length $1$ and both the units in the segment with length $2$ then you have a total of $7$ open links and $6$ closed segments, satisfying the first condition.
    \item Calculate the time considering the total number of cuts and welds. We performed $4$ cut operations and then we will need $7$ welding operations to close all the open links. The time required is $4 * 5 + 7 * 2 = 34$ minutes. This is the minimum possible time to create the necklace. 
\end{enumerate}

\subsection{Clock}
\begin{figure}[t]
  \centering
  \includegraphics[width=0.25\textwidth]{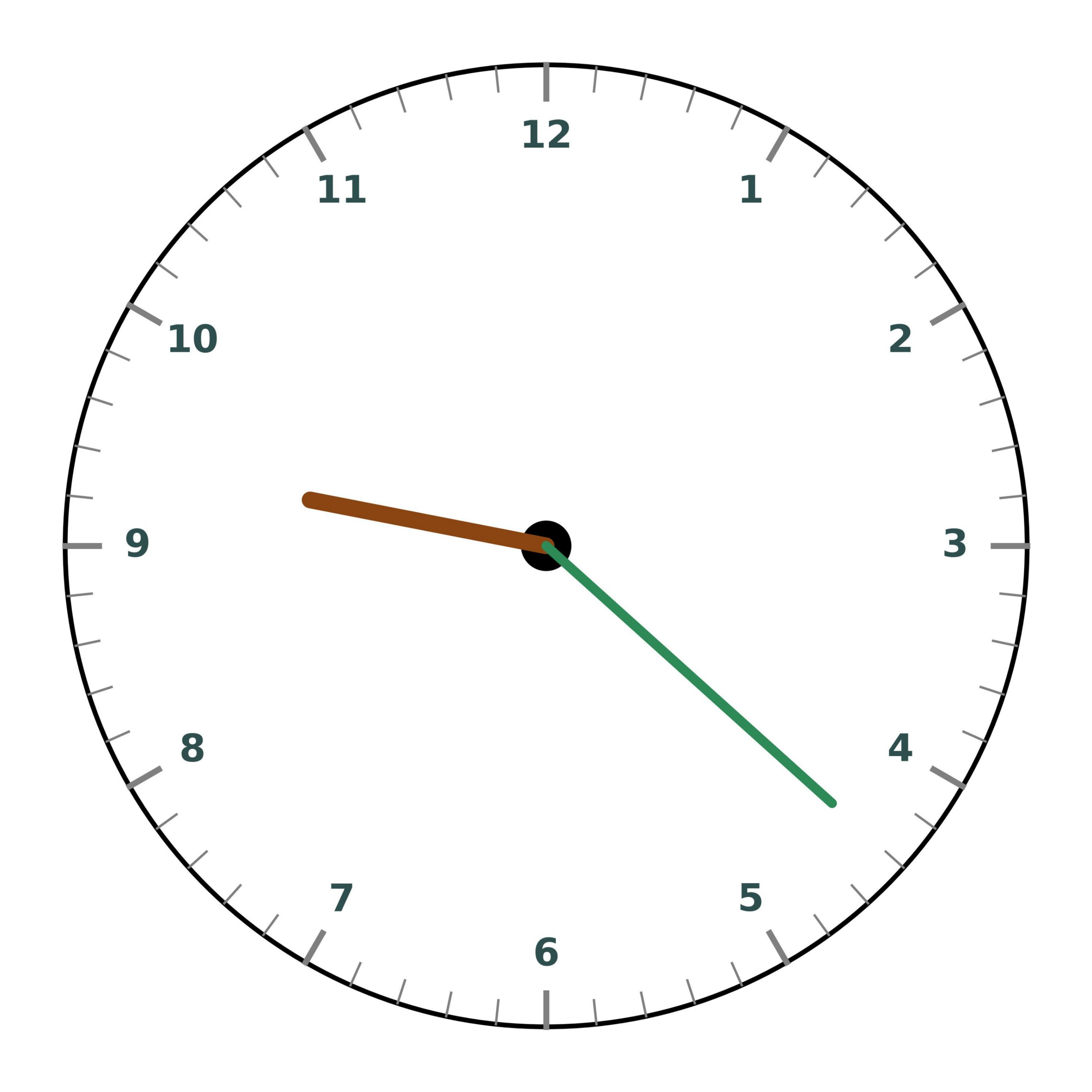}
  \caption{\footnotesize{\code{\textbf{Question:} Alexis came to an event 3 minutes ago. The current time is shown on the clock. The clock is a standard analog clock without the seconds hand. What was the time when Alexis came to the event? \textbf{Gold Answer:} 9:19}}}
  \label{fig:appendix:clock}
\end{figure}

We design an instance of a visual-temporal reasoning puzzle using clock times. We consider an analog clock with the hours, minutes hand and show a randomly chosen time as the current time. We describe an event which happened (will happen) $h$ hours and $m$ minutes ago (later). We then ask when did the event happen or when is it going to happen. We determine the answer using modular arithmetic. We show an example in \Cref{fig:appendix:clock}.

\subsection{Colour Hue}

\begin{figure}[t!]
  \centering
  \includegraphics[width=0.4\textwidth]{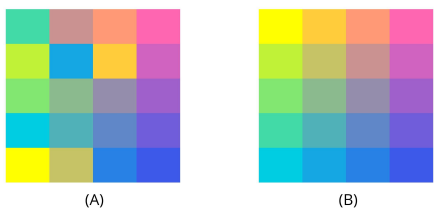}
  \caption{\footnotesize{\code{\textbf{Question:} A 5 * 4 board consists of 20 different coloured tiles. A random state of the board is shown in (A). The ideal state of the board is shown in (B). A swap consists of selecting any two tiles in the board and switching their positions. What is the minimum number of swaps required to restore the ideal state of the board from (A)? \textbf{Gold Answer:} 4}}}
  \label{fig:appendix:colour_hue}
\end{figure}

We show an example in \Cref{fig:appendix:colour_hue}. The puzzle is inspired by the game \emph{I Love Hue}\footnote{\url{https://i-love-hue.com/}}. A rectangular board contains of $m * n$ coloured tiles arranged in a rectangular grid. The board has an ideal state where the colours are arranged in a way such that each row and column shows a monotonic change in the colour shade. To create this colour arrangement, we fix the RGB colour codes of the four corner tiles and perform linear interpolation between them to determine the colour codes of the intermediate tiles. We then randomly shuffle some of the tiles to create a non-ideal arrangement of the board.
We ask how many minimum tile swaps are required to reach the ideal state from the non-ideal state. The answer can be determined using the selection sort algorithm, which minimizes the number of swaps required to sort an unsorted array. We consider the flattened version of the ideal state of the board to be the sorted state. The flattened version of the non-ideal state of the board is considered as the unsorted state. We then determine the answer using selection sort.

\subsection{Map Colouring}
\begin{figure}[h!]
  \centering
  \includegraphics[width=0.45\textwidth,height=5.5cm]{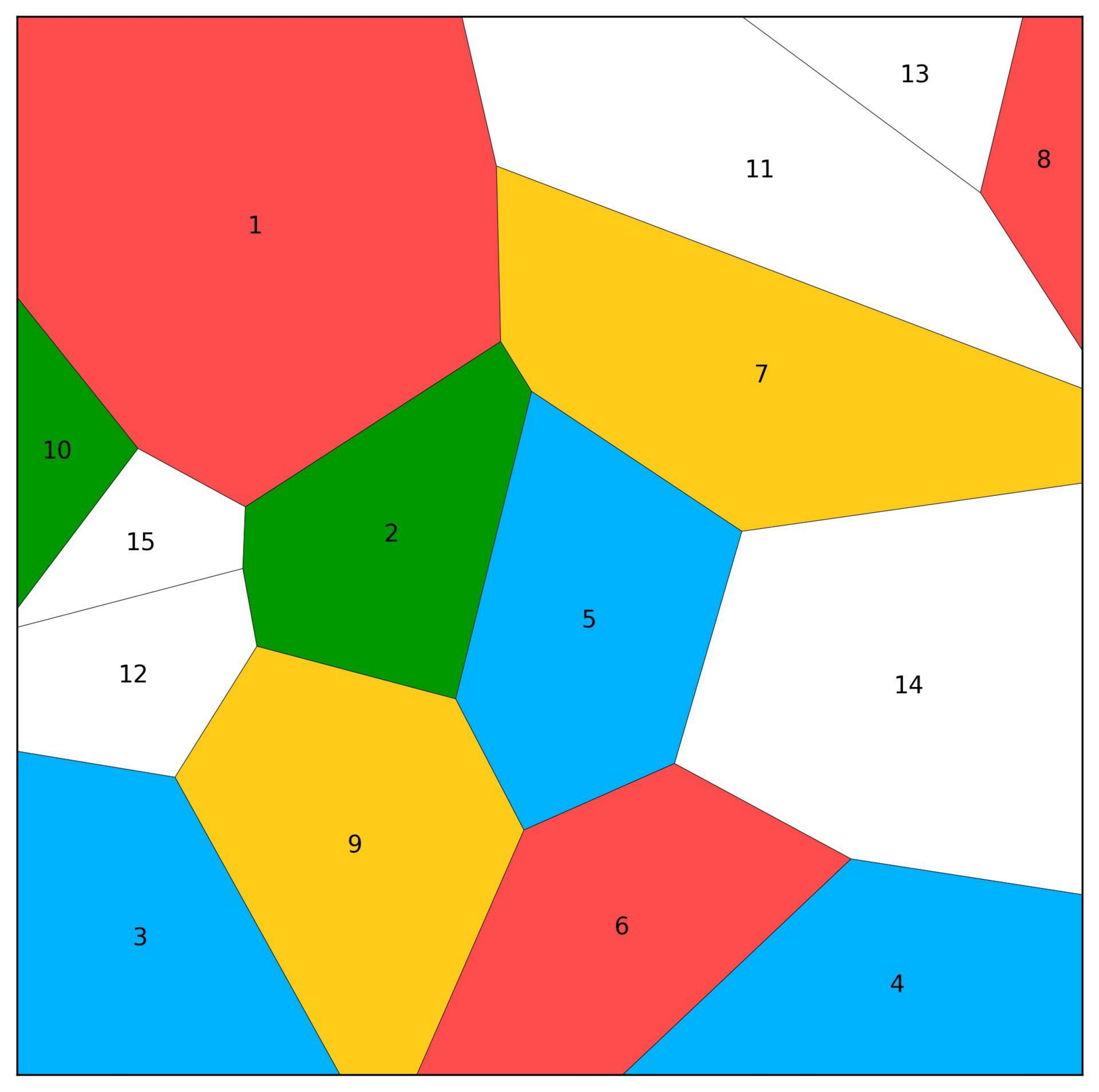}
  \caption{\footnotesize{\code{\textbf{Question:} You are given an incomplete map of a country having 15 different regions. The objective is to colour the regions of the map using only the four available colours: red, green, blue and yellow, such that no two adjacent regions have the same colour. Adjacent regions are defined as two regions that share a common boundary of non-zero length. The regions indicated by numbers 1 to 10 have already been coloured, as shown in the image. The regions indicated by numbers 11 to 15 are shown in white as they are yet to be coloured. You need to assign colours to these regions in a way such that it doesn't violate the objective. Each unique colour combination of the regions would result in a unique complete map. How many unique complete maps can be created by colouring all the white regions starting from the given incomplete map? \textbf{Gold Answer:} 8}}}
  \label{fig:appendix:map}
\end{figure}

The four colour theorem is a famous result in mathematics which states that four colours are sufficient to colour the regions of any planar map such that no two adjacent regions have the same colour. The conjecture was first proposed in the 1850s but a formal proof \citep{appel1977solution} was first developed almost 120 years later.

We create a Voronoi diagram from a finite set of points 
~\citep{Voronoi1908} followed by polygon clipping ~\citep{sutherland1974reentrant} to clip the Voronoi diagram between the finite regions of $(x, y)$, where $0 \leq x \leq 1$ and $0 \leq y \leq 1$. This region constitutes our input map. 
We represent the map as a graph with regions as nodes and their adjacent regions as the adjacency list. We use a graph colouring strategy based on Algorithm X \citep{knuth2000dancing} with four colours to find the exhaustive solutions for colouring the map. We fix the colours of some of the regions of the map and find out the number of ways the remaining regions can be coloured from the exhaustive list of solutions. We show an example of this puzzle in \Cref{fig:appendix:map}. We construct the puzzles in our dataset such that the number of masked regions is between $2$ and $6$ and the number of ways of colouring them is between $1$ and $8$. 

\subsection{Maze Solving}

\begin{figure}[t]
  \centering
  \includegraphics[width=0.3\textwidth]{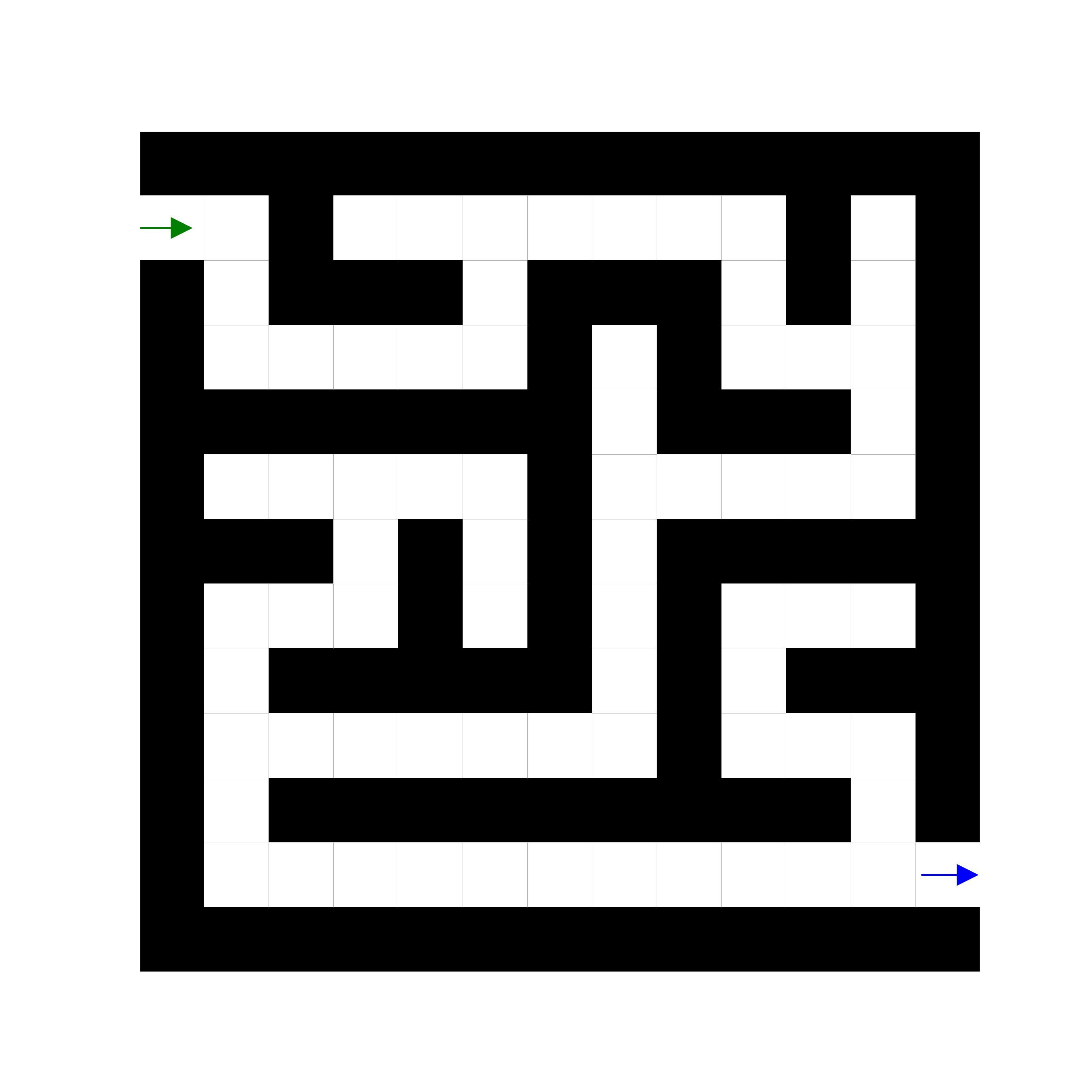}
  \caption{\footnotesize{\code{\textbf{Question:} This is maze having 13 * 13 cells. The empty cells are coloured white and the obstacle cells are coloured black. From an empty cell, you can only move up, down, left, or right to another adjacent empty cell. You cannot move diagonally between two empty cells and cannot step into a cell with an obstacle. The entry cell of the maze is shown with the green arrow. The exit cell of the maze is shown with the blue arrow. Suppose you have found the most optimal path in the maze between the entrance and exit, where you need to go through the least number of empty cells and you need to make the least number of left and right turns. What is the combined number of left and right turns do you need to make in this optimal path? \textbf{Gold Answer:} 12}}}
  \label{fig:appendix:maze}
\end{figure}

This puzzle is a typical maze path-finding problem. We start from a square/rectangular grid consisting of all black cells (walls). We define the first cell of the second row as the entrance to the maze. We then perform a directionally randomized depth-first search (DFS) with backtracking from the entrance cell to create the white cells (paths) through the maze. We also make sure that at any point of the maze, either the maximum length or the maximum width of the path is $1$ cell. This method ensures that there are no grids of white cells in the maze with both length and width greater than $2$. We finally assign the last column of the second last row or the last row of the second last column as the exit cell.

After constructing the maze, we use breadth-first search (BFS) between the entrance cell and the exit cell to find the shortest / optimal path. We then randomly select one question for this instance among the two choices: i) What is the number of left/right/total turns do you need to make in this optimal path? or ii) How many cells do you need to visit in this optimal path including the entrance and exit cells? We find out the answer to the question from the optimal path obtained from BFS. We show an example of the puzzle in \Cref{fig:appendix:maze}.

\subsection{Move Box}\label{sec:move_box}
\begin{figure}[t]
  \centering
  \includegraphics[width=0.35\textwidth]{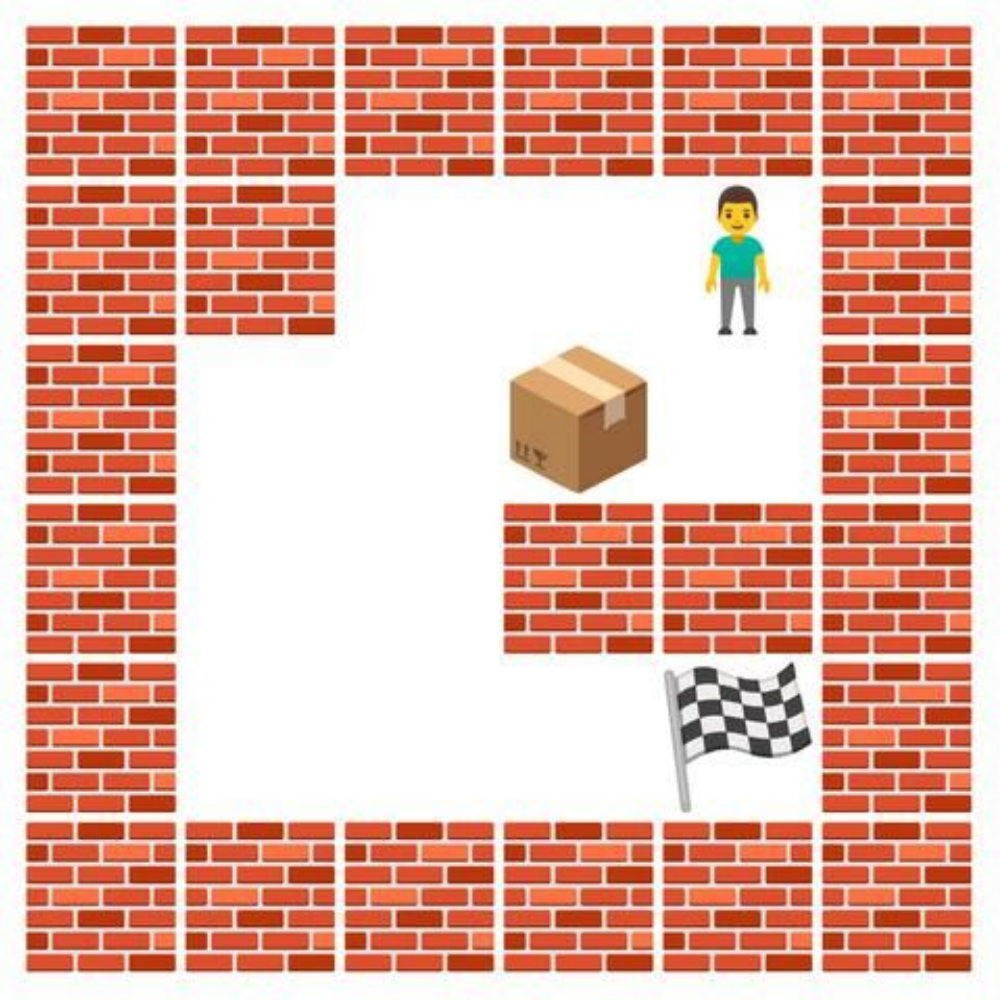}
  \caption{\footnotesize{\code{\textbf{Question:} A storekeeper is a puzzle in which the player pushes boxes around in a warehouse trying to get them to target locations. The game is represented by a 6 x 6 grid of characters grid where each element is a wall, floor, or box. Your task is to move the box to the end flag under the following rules: 1. The box can be moved to an adjacent free cell by standing next to the box and then moving in the direction of the box by 1 grid. This is a push. 2. The player cannot walk through the box. What is the minimum number of pushes to move the box to the end flag? \textbf{Gold Answer:} 5}}}\label{fig:appendix:move_box}
\end{figure}

The puzzle is inspired by a LeetCode problem \footnote{\url{https://leetcode.com/problems/minimum-moves-to-move-a-box-to-their-target-location/}}. 
The problem setting is a game in which a person pushes boxes around in a warehouse trying to get them to target locations. The objective is to move the box to the target position in the minimum number of moves. The solution can be found using breadth-first search.

\begin{figure}[t!]
  \centering
  \includegraphics[width=0.3\textwidth,height=4.75cm]{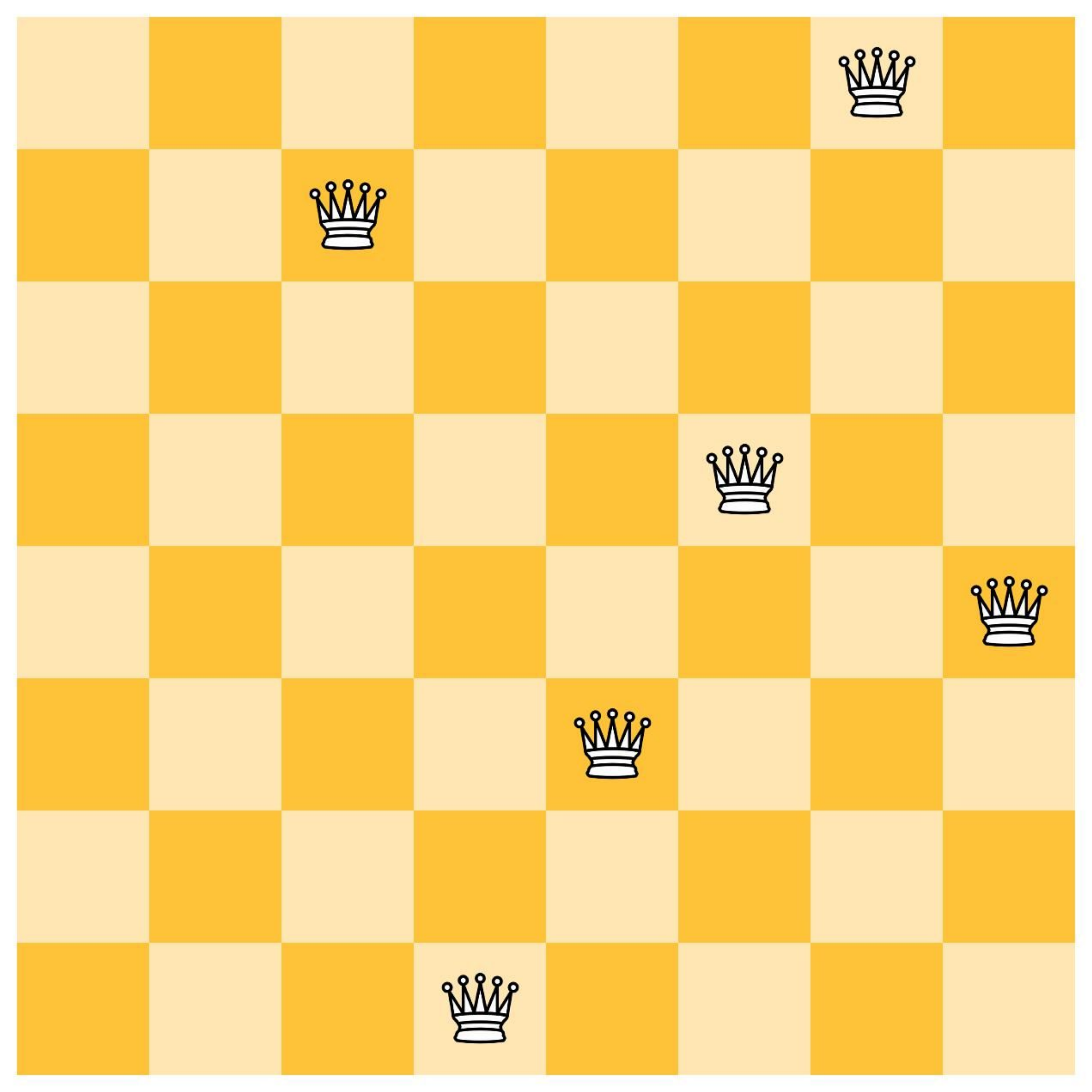}
  \caption{\footnotesize{\code{\textbf{Question:} You are given an 8 * 8 chessboard. The Manhattan distance between two squares in a chessboard is equal to the minimal number of orthogonal King moves between these squares on the otherwise empty board. The objective is to place 8 chess queens on this board so that no two queens threaten each other; i.e. no two queens share the same row, column, or diagonal. 6 queens have already been placed in some of the squares of the board, as shown in the image. Suppose you pick two squares to place the two remaining queen pieces in a way that fulfills the objective. What is the Manhattan distance between these two squares? \textbf{Gold Answer:} 5}}}
  \label{fig:appendix:n_queens}
\end{figure}

\subsection{N-Queens}

The N-Queens problem is a famous chess problem often used as an example in various computer programming techniques. The objective is to place $N$ chess queens on an $N * N$ chessboard so that no two queens threaten each other. In other words, no two queens should share the same row, column, or diagonal. We consider $N = 8, 9,$ and $10$ for which there are $92, 352,$ and $724$ solutions, respectively. We use the well-known backtracking algorithm to create the solutions to the problem. For a solution, we show the exact position of randomly chosen $N-2$ queens in the image. We ask what should be the Manhattan distance (in terms of the unit squares of the board) between the remaining $2$ queens when they are placed correctly to satisfy the objective.

The other $2$ queens can be arranged in only a single way for most cases, for which we can easily compute the Manhattan distance. In some minimal number of cases, the other $2$ queens can be placed in two different ways to satisfy the non-threatening condition in all rows, columns, and diagonals. 
However, in both of these ways, the Manhattan distance between the last $2$ queens is equal. So, we can have an exact answer to the question even though the arrangement could be distinct. We show an example of the puzzle in \Cref{fig:appendix:n_queens}.

\subsection{Number Slide}

\begin{figure}[t]
  \centering
  \includegraphics[width=0.3\textwidth]{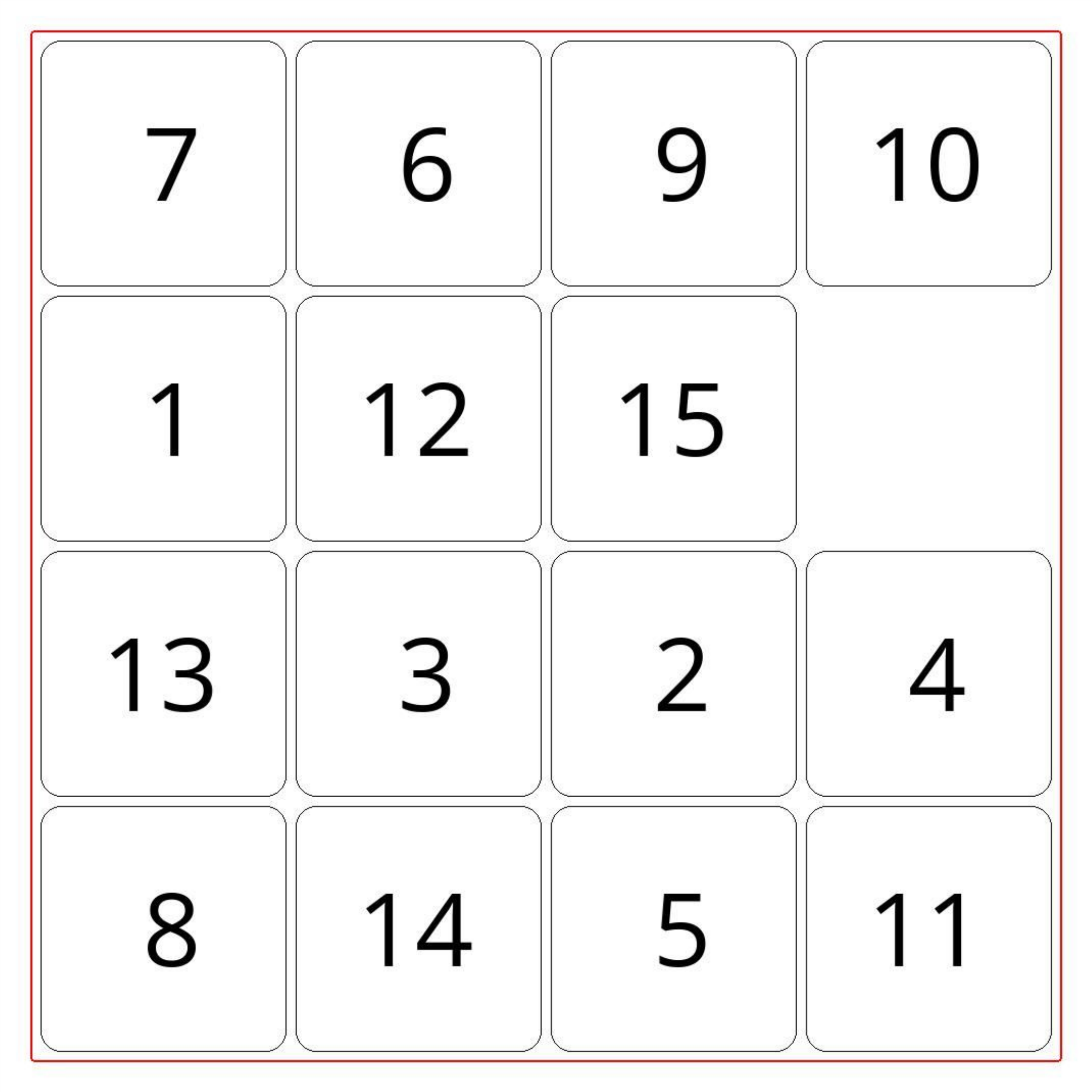}
  \caption{\footnotesize{\code{\textbf{Question:} The board shown in the image is a sliding puzzle of 4 * 4 tile dimensions. It has 15 numbered tiles and one unoccupied (open) position. Tiles in the same row or column of the open position can be moved by sliding them horizontally or vertically, respectively. All tiles always stay and move inside the red boundary wall, as shown in the image. A move is defined as moving the open position by one tile unit in any available direction. You start from the board position shown in the image and perform exactly 1 move. What is the minimum sum that you can achieve across the top most row in the final board position? \textbf{Gold Answer:} 22}}}
  \label{fig:appendix:number_slide}
\end{figure}

This puzzle is inspired by the mathematical toy known as the $15$ Puzzle\footnote{\url{https://en.wikipedia.org/wiki/15_Puzzle}}. It is a sliding puzzle board of grid size $4 * 4$, having $15$ tiles numbered 1 to 15, with one unoccupied position. Tiles can be moved by sliding them horizontally or vertically through the open position. 
A typical goal in the puzzle is to arrange the tiles in numerical order from left to right and top to bottom. We use grid sizes of $3 * 3, 4 * 4,$ or $5 * 5$ and create a random arrangement of the tiles on the board and provide it as the visual context. We then create the question in one of the following styles:

\begin{itemize}[leftmargin=4mm]
    \item How many unique board positions can be reached after performing exactly $n$ moves?
    \item What is the maximum / minimum sum that can be achieved in a  particular row / column after performing exactly $n$ moves? 
    \item You perform $n$ moves where the open position is seen to be moved in the following way: \textit{up, left, \dots}. What is the maximum / minimum / sum of numbers in the row / column that now has the open position? 
\end{itemize}

We compute the answer using breadth-first search in all the cases. We show an example of the puzzle in \Cref{fig:appendix:number_slide}.

\subsection{Rotting Fruit}

The puzzle is inspired by a LeetCode problem \footnote{\url{https://leetcode.com/problems/rotting-oranges/}}. The problem states that there is a rectangular grid with some fruits. Initially, there is a single rotten fruit. As time passes, it influences surrounding fruits to become rotten. The objective is to find out the earliest time at which all fruits become rotten. A breadth-first search algorithm can be used to find the solution.

\begin{figure}[h!]
  \centering
  \includegraphics[width=0.3\textwidth]{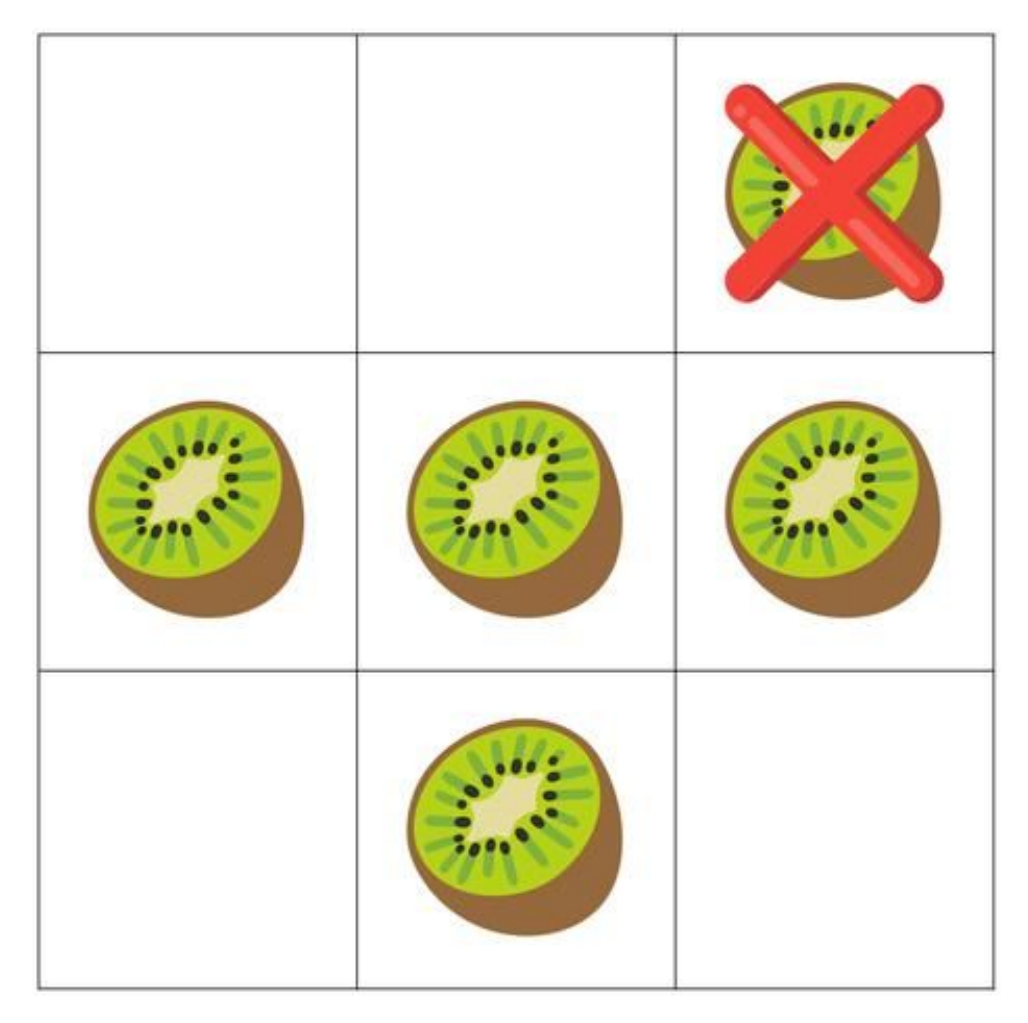}
  \caption{\footnotesize{\code{\textbf{Question:} You are given a 3 x 3 grid in which each cell can contain either no kiwi, one fresh kiwi, or one rotten kiwi. Every minute, any fresh kiwi that is 4-directionally adjacent to a rotten kiwi also becomes rotten. What is the minimum number of minutes that must elapse until no cell has a fresh kiwi? \textbf{Gold Answer:} 3}}}
  \label{fig:appendix:rotting_kiwi}
\end{figure}

\subsection{Rubik's Cube}

\begin{figure}[t]
  \centering
  \includegraphics[width=0.35\textwidth]{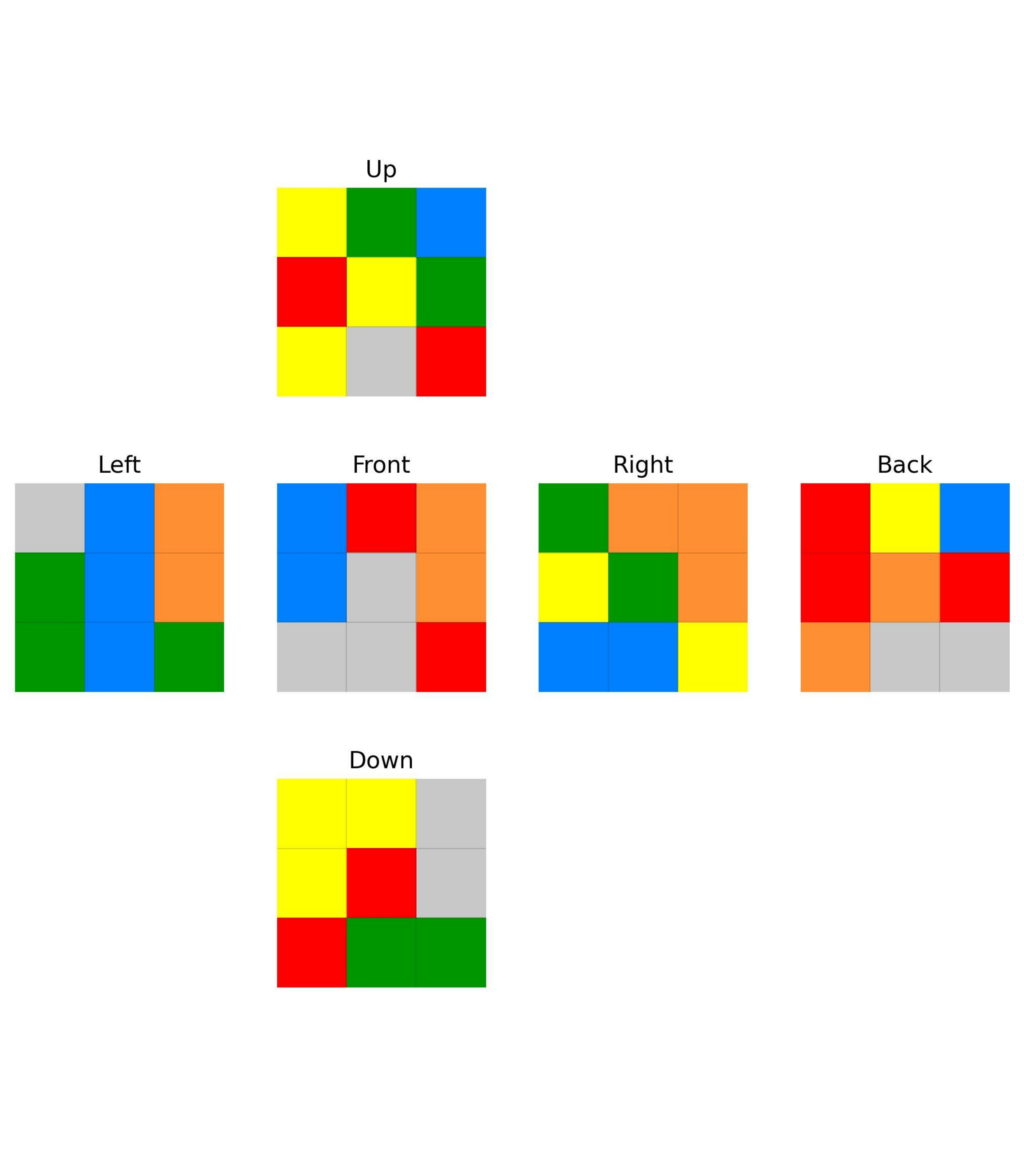}
  \caption{\footnotesize{\code{\textbf{Question:} A 3 * 3 Rubik's Cube has six different coloured panels: red, green, blue, yellow, orange, and grey. The initial state of the cube in terms of the different colour positions in its six faces is shown in the image. To represent the movements of the cube we use six letters: U for Up, D for Down, L for Left, R for Right, F for Front, B for Back. These letters are used in sequence where you need to perform each letter in the sequence from left to right. Each letter tells you to move that face clockwise by 90 degrees. A number 'n' immediately after a letter denotes that you need to move that face clockwise by 90 * n degrees. For example, `U R3' would mean rotating the up face 90 degrees clockwise and then rotating the right face 270 degrees clockwise. You perform the move sequence `D2 B' starting from the state shown in the image. What would be the number of small 1 * 1 red squares in the down face after completing the move sequence? \textbf{Gold Answer:} 2}}}
  \label{fig:appendix:rubiks_cube}
\end{figure}

Rubik's Cube is a mathematical combination toy invented by Ernő Rubik in 1974. We consider an initial state of the cube and show it as the visual context. The movements of the cube are generally denoted using the alphabets \textsc{BDFLRU} denoting clockwise movements of the back, down, front, left, right, and up faces, respectively. We first provide information about these notations in the textual context. We ask what is the number of small squares of any one of the six colours in any one of the six faces after completing a cube move sequence. The colour and the face in the question are chosen randomly for each instance. We show an example of the puzzle in \Cref{fig:appendix:rubiks_cube}.

\subsection{Think A Dot}

\begin{figure}[t]
  \centering
  \includegraphics[width=0.375\textwidth,height=5.9cm]{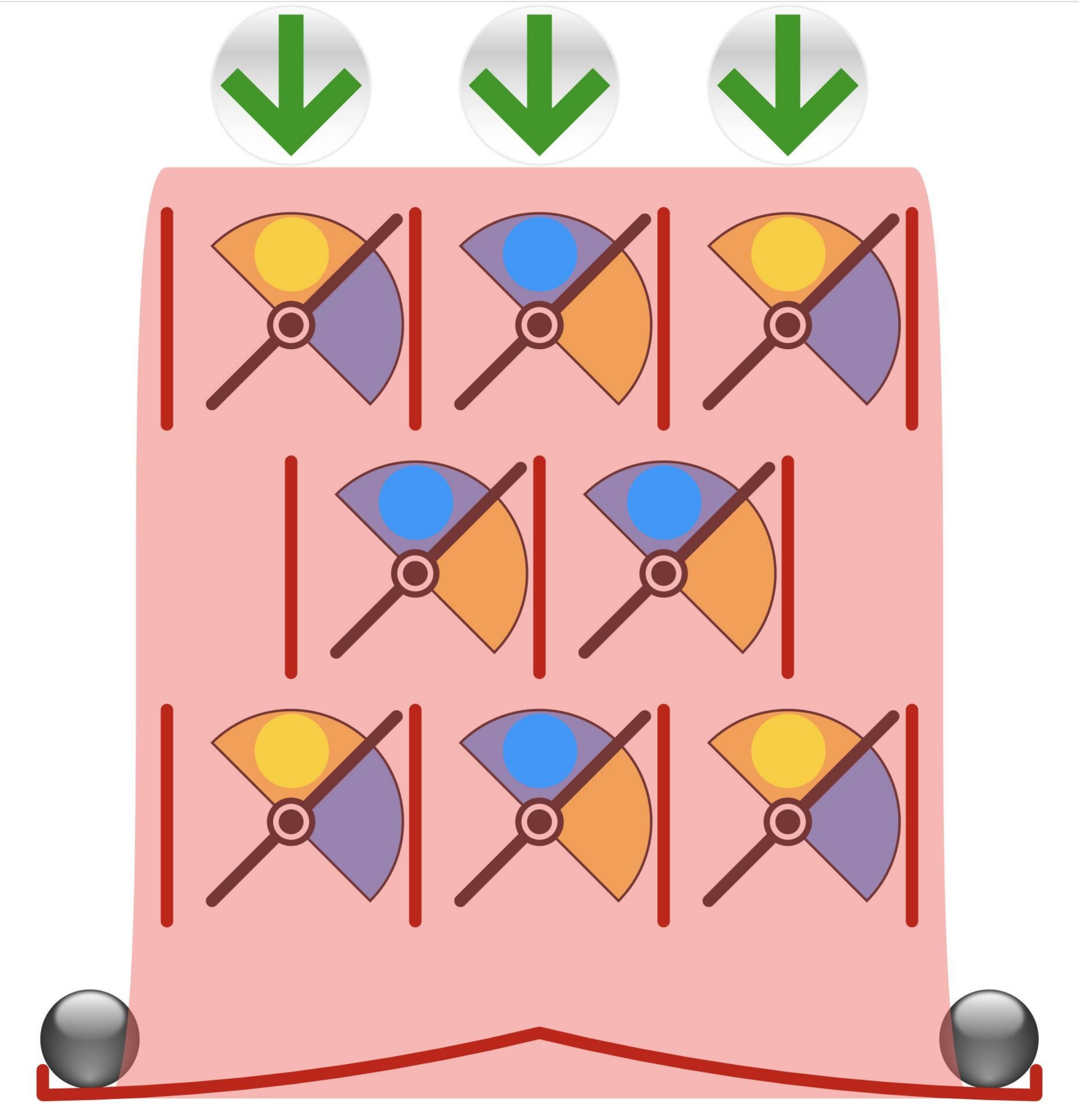}
  \caption{\footnotesize{\code{\textbf{Question:} The toy shown in the figure has eight coloured disks on its front, and three holes on its top - left, right, and center - through which a ball bearing could be dropped. Each disk would display either a yellow or blue face. When a ball passes through a disc it tips the disk mechanism which flips the face colour. The tipping of the disc mechanism determines whether the ball would be deflected to the left or to the right. The vertical walls between the discs would then determine the path of motion of the ball. A dropped ball always passes through exactly one disc in each of the top and the bottom row. Depending on the configuration of the top three discs it may or may not pass through the middle row. Finally, when the ball falls to the bottom it would exit either to a hole on the left or the right of the device. Four balls are dropped in sequence through the following holes: left, left, right, right. Consider the toy configuration after all the balls have been dropped and they have exited from the bottom. How many yellow faces can be seen in total in all the rows now? \textbf{Gold Answer:} 6}}}
  \label{fig:appendix:think_dot}
\end{figure}

Think-a-Dot is a mathematical toy invented by Joseph Weisbecker\footnote{\url{https://en.wikipedia.org/wiki/Think-a-Dot}}. 
It has three holes on its top through which a ball bearing could be dropped. It also has and eight coloured disks each displaying a blue or yellow face. 
When a ball is dropped through the toy, it would flip the disk mechanisms that it passed, and it would determine whether the ball would be deflected to the left or the right. 
We show an example of the puzzle in \Cref{fig:appendix:think_dot}. We start with an initial configuration of the toy with a specific combination of the disk colours. We choose between $1$ to $4$ balls and a sequence of dropping them between the left, center and right holes. We ask how many blue / yellow disk faces can be seen in the top row / middle row / bottom rows / all rows after dropping the balls in that sequence. We write an algorithm to determine the answer from the initial configuration and the sequence of drops considering the state of the rows, the flipped disks and the position of the walls. 
 
\subsection{Tower of Hanoi}

\begin{figure}[t]
  \centering
  \includegraphics[width=0.4\textwidth]{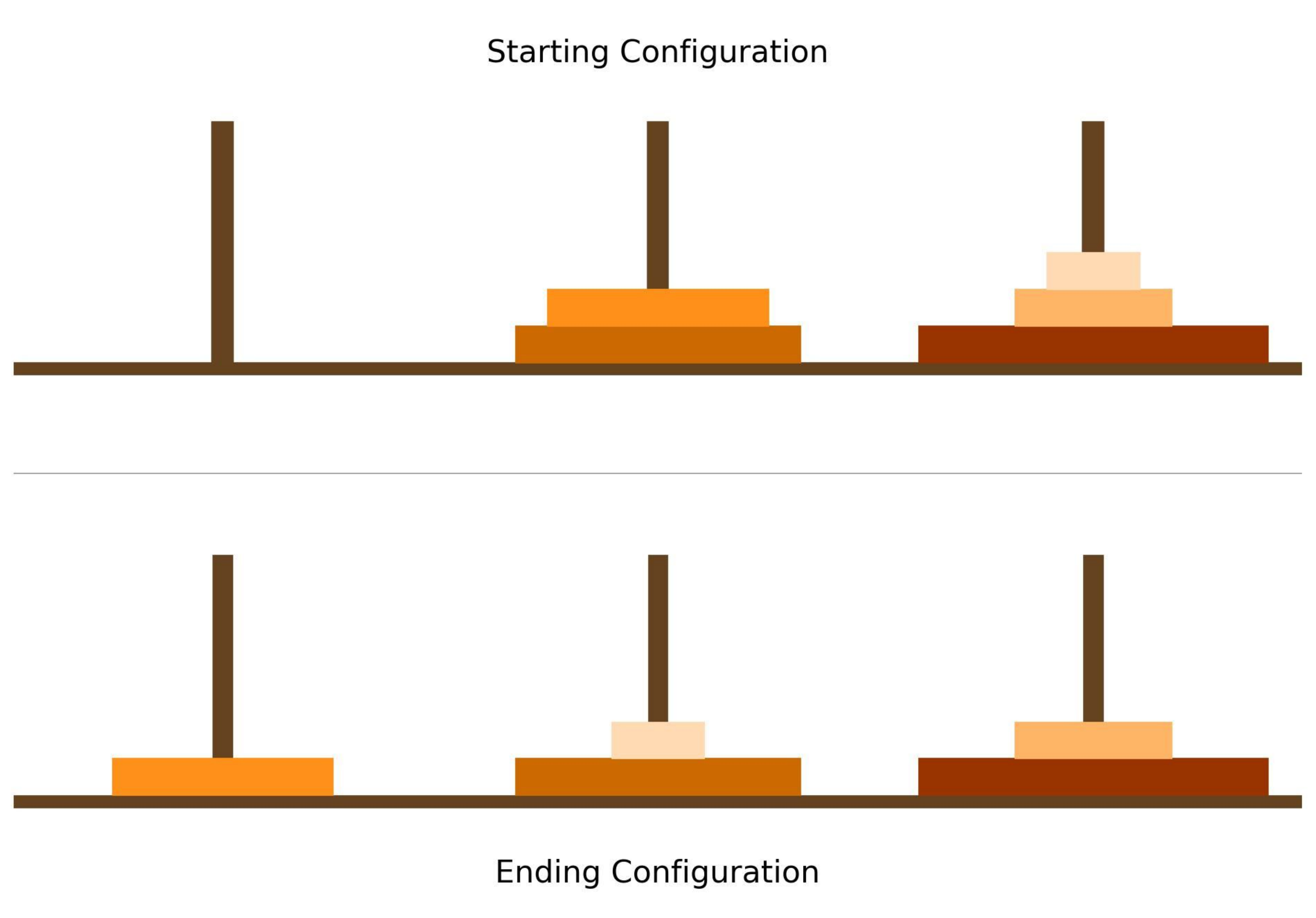}
  \caption{\footnotesize{\code{\textbf{Question:} You are playing a Tower of Hanoi game with 3 rods and 5 disks of various diameters, which can slide onto any rod. You are given the starting and ending configuration of the game as shown in the top and the bottom of the image, respectively. The game has the following rules: i) Only one disk may be moved at a time; ii) Each move consists of taking the upper disk from one of the stacks and placing it on top of another stack or on an empty rod; and iii) No disk can be placed on top of a disk that is smaller than it. What is the minimum number of moves required to go from the starting to the ending configuration? \textbf{Gold Answer:} 2}}}
  \label{fig:appendix:tower_of_hanoi}
\end{figure}

The Tower of Hanoi is a well-known mathematical game often used in teaching the fundamentals of computer programming. The game consists of $3$ rods and $n$ disks of various diameters, that can slide into any rod. In the original version of the puzzle, we start with all the disks stacked on one rod in order of decreasing size. The goal is to move the full stack of disks to another. Moving the discs between the rods is constrained by the following rules: (i) We can only move one disc at a time, (ii) Each move consists of taking the upper disk from one of the stacks and placing it on top of another stack or on an empty rod, (iii) No disk can be placed on top of a disk that is smaller than it. With no other constraints, the original version of the puzzle can be solved in a minimum of $2^n - 1$ moves. 

We consider the number of discs to be between $3$ to $6$ and generate the optimal solution considering the original problem definition. We then select two random configurations of the game from the optimal solution, which are at most $k = 6$ moves away from each other. We consider them to be the starting and the ending configuration. 
As the original solution is optimal, the sequence of moves required to reach the ending configuration from the starting configuration is also optimal. 
We ask what is the minimum number of moves required to reach the ending configuration from the starting configuration. We mark the answer to be $k$. 
We show an example of the puzzle in \Cref{fig:appendix:tower_of_hanoi}.

\subsection{Water Jugs}

\begin{figure}[t!]
  \centering
  \includegraphics[width=0.4\textwidth,height=6cm]{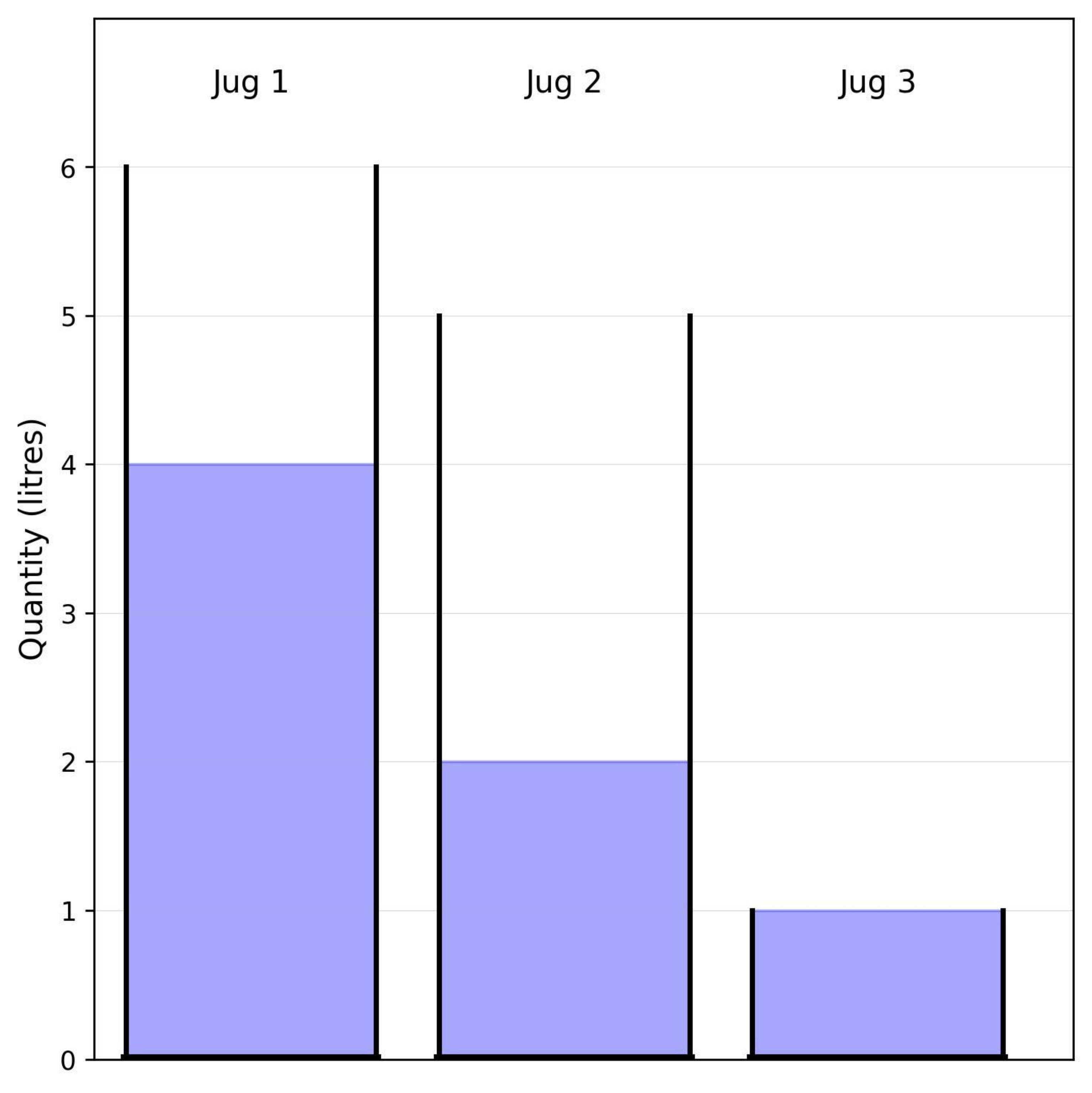}
  \caption{\footnotesize{\code{\textbf{Question:} You are given 3 jugs of capacities 6, 5, 1 litres. Initially, the amount of water that is contained in each jar is shown in the image. A single step of water pouring from one jug to another is constrained by the following rules: i) take a non-empty jug and pour water from it to another non-full jug until the first one becomes empty or the second one becomes full, and ii) no water can be spilt while pouring. The objective is to reach the amounts of 4, 3, 0 litres of water in the jugs from left to right, respectively. What is the minimum number of water pouring steps required to achieve the objective? \textbf{Gold Answer:} 1}}}
  \label{fig:appendix:water_jugs}
\end{figure}

This puzzle belongs to a class of measuring puzzles involving a finite collection of water jugs with integer capacities. We provide the initial amount of water present in each jug as the visual context. We then ask how many steps of water pouring are required to reach a goal state defined in terms of specific quantities of water present in each jug. The water pouring steps are constrained by a couple of rules: i) water can be poured from a non-empty jug to another non-full jug until the first one becomes empty or the second one becomes full, and ii) no water can be split during pouring. 

We consider the number of jugs to be between $3$ and $5$, with each initially having an amount between $1$ and $14$ litres of water. We create a pool of random goal states having the same quantity of total water with each jug having water that is less or equal to its respective capacity. We only consider goal states that are reachable from the initial state using breadth-first search. We show an example of the puzzle in \Cref{fig:appendix:water_jugs}.

\subsection{Wheel of Fortune}

\begin{figure}[t]
  \centering
  \includegraphics[width=0.35\textwidth]{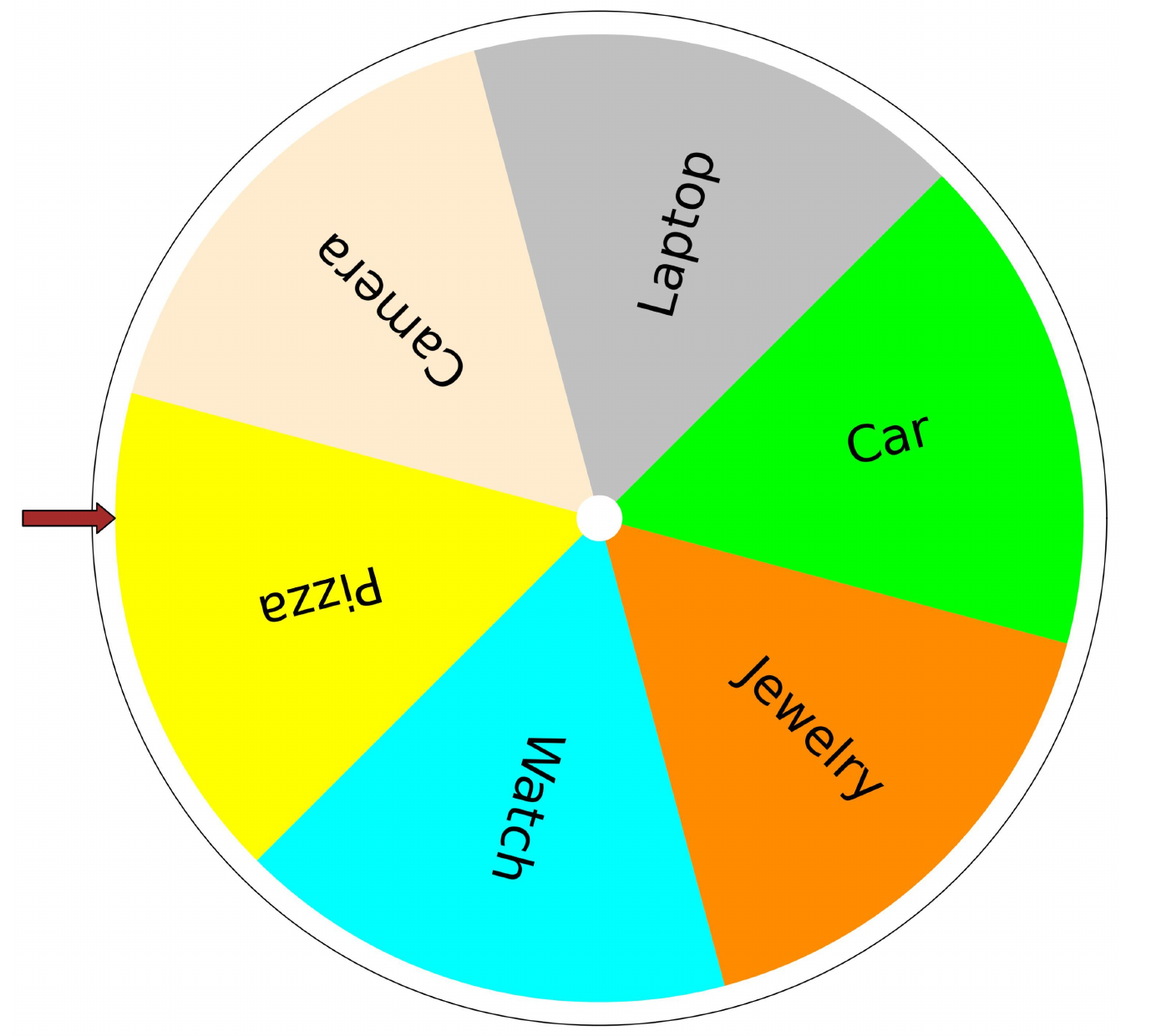}
  \caption{\footnotesize{\code{\textbf{Question:} A fortune wheel has 6 segments of different colour. The initial position of the wheel is shown in the figure. Each segment is associated with a prize as shown in the embedded text within the segment. The axis of rotation of the wheel passes through its center and is perpendicular to the surface of the wheel. You spin the wheel clockwise and it rotates 1695 degrees before stopping. You are going to win the prize for the segment that now falls in front of the brown arrow. What is your prize? \textbf{Gold Answer:} Laptop}}}
  \label{fig:appendix:wheel_of_fortune}
\end{figure}

We design this spinning wheel puzzle to assess the spatial reasoning ability of foundation models. We sketch a wheel with $6,  8$ or $10$ segments with different colours and associate each of them with a prize. The angular span of the segments is chosen to be either uniform or random. We show the initial position of the wheel and the position of a fixed arrow as the visual context. We ask what the prize would be (from the segment in front of the arrow) after the wheel has been rotated by a certain amount of degrees or full rotations in clockwise / anti-clockwise direction. We determine the answer using simple rotational mechanics. We show an example of the puzzle in \Cref{fig:appendix:wheel_of_fortune}.

\subsection{Wood Slide}

\begin{figure}[t]
  \centering
  \includegraphics[width=0.35\textwidth]{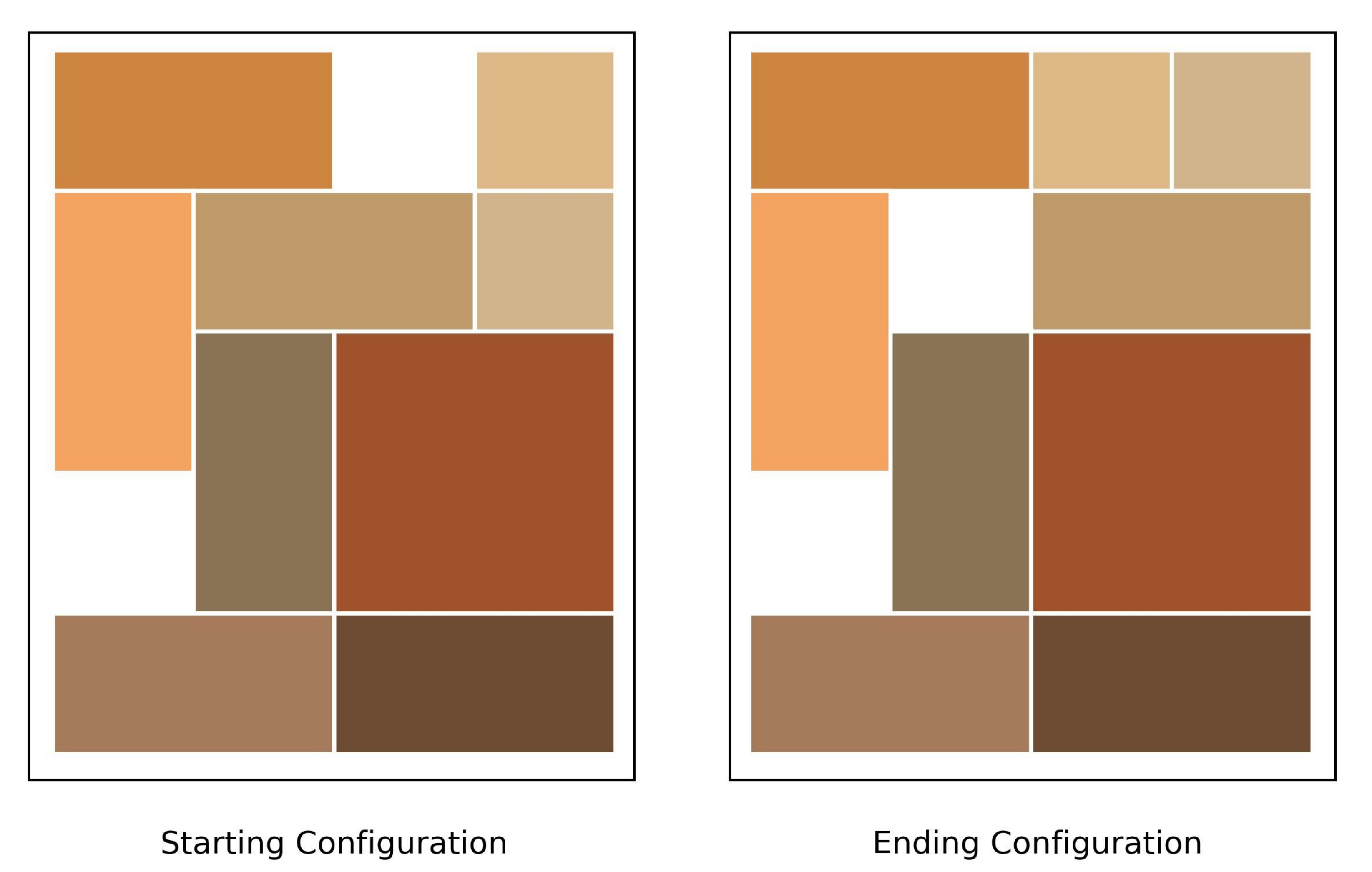}
  \caption{\footnotesize{\code{\textbf{Question:} Consider a sliding block puzzle of grid size 5 * 4 units. It has 9 wooden blocks of varying sizes: one 2 * 2, four 1 * 2, two 2 * 1, and two 1 * 1. The gird also has two empty 1 * 1 spaces. The blocks cannot be removed from the grid, and may only be slid horizontally and vertically within its boundary. A move is defined as selecting a block that is slideable, and moving it by 1 unit either horizontally or vertically, whichever is possible. The image shows the starting and ending configurations of the puzzle grid. The wooden blocks are shown in various shades of brown and the empty spaces are shown in white. What is the minimum number of moves required to reach the ending configuration from the starting configuration? \textbf{Gold Answer:} 3}}}
  \label{fig:appendix:wood_slide}
\end{figure}

This puzzle is inspired by the Klotski\footnote{\url{https://en.wikipedia.org/wiki/Klotski}} sliding puzzle.
We consider a puzzle grid of size 5 * 4 units that has 9 wooden blocks of various sizes: one 2 * 2, four 1 * 2, two 2 * 1, and two 1 * 1. The other two spaces are empty. There is a version of the Klotski puzzle known as the Pennant Puzzle where we start with the largest 2 * 2 block residing at the top left. The objective is to bring this piece to the bottom left by sliding the available pieces. The shortest solution of this puzzle consists of 83 moves. We first use breadth-first search to find this optimal solution. For each instance in our dataset, we choose two board positions encountered in this solution such that they are at most 5 moves away from each other. We consider these positions as the starting and ending configuration of the board. We then create the question that asks the minimum number of moves required to reach the ending configuration from the starting configuration. We show an example of the puzzle in \Cref{fig:appendix:wood_slide}.

\section{Negative Choice Generation for MCQA}
\label{sec:appendix:mcqa}
The negative choices are constrained to \textit{Yes} and \textit{No} for \textit{Board Tiling}. The negative choices are constrained to the prizes that appear in the wheel for \textit{Wheel of Fortune}. All other puzzles in \dataset{} have numerical answers. We use the heuristics of randomly sampling numbers within the same magnitude as the gold answer to create the negative choices for all the other puzzles. For example, if the gold answer is less or equal to 6 then we choose negatives between 1  - 6; otherwise, if the gold answer is less or equal to 10 then we choose negatives between 1  - 10. We step up the ranges to 50 and 100 next.

\section{Model Configurations}
\paragraph{GPT-4V} We use the publicly available API to query the \code{gpt-4-vision-preview} version of the model.
\paragraph{Gemini Pro} We use the publicly available API to query the \code{gemini-pro-vision} version of the model.

\end{document}